\documentclass{article}

\usepackage{jmlr2e}
\usepackage{systeme}
\usepackage{amsmath}

\usepackage[utf8]{inputenc} 
\usepackage[T1]{fontenc}    
\usepackage{hyperref}       
\usepackage{url}            
\usepackage{booktabs}       
\usepackage{amsfonts}       
\usepackage{nicefrac}       
\usepackage{microtype}      
\usepackage{xcolor}         
\usepackage{algorithm}
\usepackage{algpseudocode}
\usepackage{amsmath,amssymb,bm,commath}
\usepackage{graphicx}
\usepackage{dsfont}

\title{Decomposed Linear Dynamical Systems (dLDS) for \\  learning the latent components of neural dynamics}

\author{%
  \name Noga Mudrik\thanks{Equal Contribution} \email nmudrik1@jhu.edu\\
  \addr Department of Biomedical Engineering \\ Center for Imaging Science\\ Kavli NDI \\ 
  Johns Hopkins University\\
  Baltimore, MD 21218. \\
   \AND
  \name Yenho Chen\footnotemark[1] \email yenho@gatech.edu \\
\addr Department of Biomedical Engineering \\
  Georgia Institute of Technology \\
  Atlanta, GA 30332. \\
  \AND
  \name Eva Yezerets \email eyezere1@jhu.edu\\
  \addr Department of Biomedical Engineering \\ Center for Imaging Science\\
  Johns Hopkins University\\
  Baltimore, MD 21218. \\
  \AND
  \name Christopher J. Rozell \email crozell@gatech.edu\\
  \addr School of Electrical and Computer Engineering \\
  Georgia Institute of Technology \\
  Atlanta, GA 30332. \\
  \AND
  \name Adam S. Charles 
  \email adamsc@jhu.edu \\
  \addr Department of Biomedical Engineering \\
  Center for Imaging Science \\
  Mathematical Institute for Data Science \\ 
  Kavli NDI\\
  Johns Hopkins University \\
  Baltimore, MD 21218. \\
}

\editor{TBD}
\date{May 2023}

\begin{document}
\maketitle

\begin{abstract}
    Learning interpretable representations of neural dynamics at a population level is a crucial first step to understanding how observed neural activity relates to perception and behavior. Models of neural dynamics often focus on either low-dimensional projections of neural activity, or on learning dynamical systems that explicitly relate to the neural state over time. We discuss how these two approaches are interrelated by considering dynamical systems as representative of flows on a low-dimensional manifold. Building on this concept, we propose a new decomposed dynamical system model that represents complex non-stationary and nonlinear dynamics of time series data as a sparse combination of simpler, more interpretable components. Our model is trained through a dictionary learning procedure, where we leverage recent results in tracking sparse vectors over time.  The decomposed nature of the dynamics is more expressive than previous switched approaches for a given number of parameters and enables modeling of overlapping and non-stationary dynamics. In both continuous-time and discrete-time instructional examples, we demonstrate that our model can well approximate the original system, learn efficient representations, and capture smooth transitions between dynamical modes, focusing on intuitive low-dimensional non-stationary linear and nonlinear systems. Furthermore, we highlight our model’s ability to efficiently capture and demix population dynamics generated from multiple independent subnetworks, a task that is computationally impractical for switched models. Finally, we apply our model to neural ``full brain'' recordings of \textit{C. elegans} data, illustrating a diversity of dynamics that is obscured when classified into discrete states.
\end{abstract}

\section{Introduction}

The past decade has seen rapid growth in neuroscience driven by the emergence of new technologies that enable the recording of neural population activity, such as large-scale optical imaging~\citep{demas2021high} and electrophysiology~\citep{steinmetz2021neuropixels}. As a result, neural data analysis has moved beyond the characterization of single neurons to the modeling of entire large neuronal populations~\citep{saxena2019towards}. For large simultaneous recordings, a significant challenge lies in understanding the intricate correlated patterns of neural activity on a single trial basis. Currently, the quantitative language describing these observations has primarily leveraged two conceptual frameworks: 1) dimensionality reduction and 2) dynamical systems modeling. 

Dimensionality reduction primarily addresses the identification of a small number of degrees of freedom that characterize a time series recording. Often these methods treat activity patterns at individual time points independently and seek to discover a representative geometry that underlies the data. Traditional linear methods, including PCA, ICA, and their variants~\citep{yu2008gaussian,wu2017gaussian} have recently given way to more flexible descriptions of dimensionality reduction. 
In particular, the \emph{neural manifold} assumption, i.e., that instantaneous neural activity patterns lie on a low-dimensional manifold, removes the assumption of linearity in the low-dimensional neural representation, and enables the identification of correlated activity that corresponds to a potentially much lower-dimensional geometric structure. 
Many recent nonlinear dimensionality reduction methods rely on the manifold hypothesis~\citep{wu2017gaussian,nieh2021geometry,gallego2017neural,cunningham2014dimensionality,Benisty2021.08.15.456390,7548324}, including local embeddings~\citep{balasubramanian2002isomap,roweis2000nonlinear} and variational auto-encoders~\citep{han2019variational}. These methods, however, often fail to fully capture the temporal nature of the data.

By contrast, dynamical systems models focus primarily on capturing the temporal relationships within neural activity. However, these models often treat activity patterns over time as arising from individual latent states without regard to any geometric structure being modeled by these low-dimensional latent variables. Proposed dynamical systems models include concise linear dynamical systems (LDS)~\citep{churchland2012neural}, switched systems that capture abrupt non-stationarities ~\citep{linderman2017bayesian,glaser2020recurrent,nassar2018tree}, GLMs that learn sets of filters to describe temporal conditional probabilities~\citep{pillow2008spatio}, and more recently arbitrary function approximations in the form of general recurrent neural networks (RNNs)~\citep{pandarinath2018inferring,keshtkaran2019enabling}. Both approaches contain a fundamental trade-off between model complexity and interpretability. On one side are simple, regularized, often linear models that directly expose latent relationships in the data, but are limited in their ability to accurately model complex structures. On the other side are modern black-box deep learning methods, which are highly expressive but obfuscate learned relationships and are thus difficult to interpret~\citep{schulz2020different}. While these models remain a promising avenue for capturing the temporal dynamics of neural activity, they often lack explicit constraints on the structure of the latent variables or on the underlying neural manifold.  

Accordingly, there is a critical need to develop methods that remain interpretable while maintaining a high level of expressivity, i.e., capturing rich nonlinear structure that arises when studying neural population activity. While both traditional dimensionality reduction methods and dynamical systems models seek to represent low-dimensional geometric structures, we currently lack methods that integrate the manifold hypothesis directly into a dynamical systems model, thus maintaining both model capacity and interpretability. This gap in modeling limits our ability to study important problems in neuroscience. As a prime example, one task that remains challenging is the identification and characterization of multiple subnetworks within a neural population. Population recordings often consist of distinct subnetworks that exhibit different functional roles or represent different aspects of information processing. Understanding the dynamics of each distinct subnetwork is crucial for revealing the underlying mechanisms of neural computation. While each subnetwork may have its own dynamics, connectivity patterns, functional properties, and interactions with other subnetworks, the multivariate time series in neural recordings represent all subnetworks' combined activity and must be unmixed to discover the true underlying structure.  

To address this need, we introduce a decomposed Linear Dynamical Systems (dLDS) model that describes high-dimensional neural activity as dynamical flows on a low-dimensional manifold. Transitions between consecutive time points are decomposed as a time-varying mixture of linear dynamics systems (LDSs). In our approach, each LDS element captures a canonical movement along the neural manifold, and are linearly combined to model the overall dynamics in the low-dimensional data geometry. Specifically, we constrain the linear combination at each time point to be sparse, i.e., only a few dynamics are combined at any time point. This parsimony eases the task of interpreting the individual LDS systems in light of global dynamical structure or external variables. dLDS exhibits increased expressiveness per model parameter than existing switched Linear Dynamical Systems (SLDS), and can capture richer dynamics while maintaining interpretability, as it uses simple LDS primitives that highlight unique dynamical patterns under different conditions. 

Our implementation of dLDS encompasses both continuous- and discrete-time variants, each with distinct advantages under different settings, such as scalability and invariance to system speed, respectively. Our main contributions in this paper are 1) introducing the dLDS model to illustrate its foundational properties, capabilities, and strengths; 2) demonstrating key advantages of this model over current approaches in multiple scenarios; and 3) illustrating its use in the analysis of population neural activity from calcium imaging in \textit{C. elegans}.

\section{Background and Related Work}

\textbf{Dynamical systems models of neural data.}
Dynamical systems models explicitly seek to identify the relationships of neural activity patterns over time. Early work leverages the learning of LDSs to approximate the progression of activity in neural state space~\citep{sani2021modeling,golub2013learning,churchland2012neural}. These approaches leverage theory from traditional linear stationary system identification literature, such as Kalman filtering, etc.~\citep{Haykin:1996}. Applications include characterization of neural populations, controlling of brain-computer interfaces, and relating brain activity to external variables~\citep{yu2007mixture}.
Neural activity, however, is characterized by nonlinear and non-stationary dynamics, and two primary approaches have been taken so far for addressing this challenge: 1) nonlinear recurrent neural networks~\citep{pandarinath2018inferring,keshtkaran2019enabling,sussillo2015neural,sani2021all,kleinman2021mechanistic}, and 2) Switched Linear Dynamical Systems (SLDS)~\citep{Ackerson1970, Chang1978,Hamilton1990,Bar-Shalom1993,Ghahramani96switchingstate-space,murphy1998switching,NIPS2008_950a4152,linderman2017bayesian}. Recurrent neural network (RNN) approaches, while highly flexible, are based on training ``black boxes'' which make interpretability of the learned dynamics difficult. Moreover, some emerging RNN-based approaches, e.g.,~\citep{pandarinath2018inferring,keshtkaran2019enabling}, compress the dynamics in unintuitive ways, such as learning initial conditions that recapitulate the neural dynamics when input into the trained RNN. 

SLDS, on the other hand, explicitly seeks interpretable characterizations of the dynamics by modeling transitions over time between a discrete set of linear systems. SLDS variants are described as a Gauss-Markov process and a switching model that determines transitions between the linear dynamics through a discrete-time Markov process. Although SLDS can discover latent dynamics, it is a limited generative model since state durations are determined stochastically. Recurrent SLDS (rSLDS) extends SLDS by including an additional dependency between the discrete switches and the previous location in state space through a stick-breaking logistic function~\citep{linderman2017bayesian}. rSLDS improves interpretability by dividing the state space into $K$ partitions with locally linear dynamics. However, rSLDS introduces problems during inference due to its dependencies on the permutation of the discrete switches. Tree-structured recurrent SLDS is an extension of rSLDS that addresses this issue through a generalized stick-breaking procedure that enables efficient inference and the ability to represent dynamics at multiple levels of resolution~\citep{nassar2018tree}. However, scalability remains an issue as the optimal number of discrete systems is difficult to determine, often leading to inefficient representations of the underlying dynamics. 

We note that all prior approaches treat the system as a single dynamical system, with potential changes over time. However these approaches all fail to address the required capability to identify subsystems that overlap in time, obscuring, e.g., slow-timescale components when faster-timescale systems are simultaneously observed. To address this drawback we take inspiration from other data representation models, in particular sparse coding. 

\textbf{Sparse coding and dictionary learning.}
Learning data representations is a central theme in machine learning and neuroscience. One fundamental approach is sparse coding, which assumes a form of efficiency in representation under a linear generative model~\citep{OLS:1996,aharon2006rm}. In the sparse coding  representation, each data point $\bm{y}_k$ can be linearly generated from a latent vector $\bm{a}_k$ such that $\bm{y}_k = \bm{D}\bm{a}_k + \bm{\epsilon}$, where the matrix $\bm{D}$ contains representational features as its elements, and $\bm{\epsilon}$ is representational noise. Sparse coding assumes that the representation is efficient in that for any $\bm{y}_k$, only a few of the features (columns of $\bm{D}$) are required to construct the data point, i.e., the number of non-zero entries in $\bm{a}_k$ for any $\bm{y}_k$ is assumed to be small relative to the size of $\bm{a}_k$. In general, $\bm{D}$ is unknown and must be learned from data via a process called dictionary learning (DL). DL can be expressed as a variational approach with a delta approximation to the posterior~\citep{barello2018sparse} results in a two-step optimization that iterates between inferring the sparse vector $\bm{a}_k$ for a subset of points, and updating the dictionary $\bm{D}$ by taking a gradient step over the approximated likelihood.

\textbf{Transport Operators.} 
Related to dLDS is the Transport Operators (TOs) framework \citep{culpepper2009TOs}. TOs represent a class of generative manifold models by characterizing transformations in local regions as a continuous-time-evolving LDS model. The transformation matrix is decomposed as a weighted sum of $K$ dictionary elements called \emph{TOs}, each individually representing movement along a particular path on the latent manifold and can be combined to reconstruct observed transformations. Operators are learned as part of the DL iterative procedure and are updated at each iteration by taking a gradient descent step to minimize the desired cost. This approach has been demonstrated to effectively approximate geometrical structures of nonlinear manifolds in a variety of settings. For complex data, TOs have been integrated into the latent space of autoencoders and VAEs which enables generative transformation paths and meaningful extrapolations~\citep{Connor2020mae,connor2021vaells}. More recent advances include relaxing the requirement of transformation labels, learning local operator statistics for identity-preserving transformations, and improving the scalability of inference for higher-order models~\citep{Connor2021identitypreserving, fallah2022variational}. Despite this progress, existing work on TOs mainly focuses on discovering the geometry of latent manifolds, but fails to describe the nonlinear temporal flows on these manifolds in complex time series data. 

\textbf{Interpretability.}  
Despite the myriad of existing definitions, interpretability is an important consideration to understanding neural dynamics. Here, we discuss our approach to defining interpretability. \cite{montavon2018methods} states, ``Interpretations map an abstract concept into a domain that a human can make sense of. In the context of neural signals, data is collected from arbitrary nonlinear dynamical systems which generally have no analytical solutions available, making it challenging to predict and interpret their behavior. To facilitate the ease of human reasoning, a model must be transparent---both with respect to the whole model altogether as well as for each of its individual parts separately~\citep{lipton2018mythos}. Reasoning about the entire model can be achieved through simple, parsimonious representations, while reasoning about its components is accomplished through statistically transparent structures that can be easily modified and analyzed. This suggests that an interpretable model is one that is sparse with simple building blocks. 

dLDS takes inspiration from these guidelines and achieves transparency and parsimony through three separate mechanisms. First, dimensionality reduction finds a latent low-dimensional representation that summarizes and approximately generates the correlated high-dimensional signals. Second, the latent nonlinear system is locally approximated with linear systems. Linear components are fully characterized mathematically, making it possible to determine the behavior of the system through its eigendecomposition and gain insight into the underlying mechanisms at any point in time. Third, each transition is decomposed into a sparse combination of a dictionary of linear dynamics which improves parsimony through the reuse of the learned dictionary to efficiently represent shared dynamics over time and under different conditions.

\textbf{Expressivity.}
Expressivity is the ability for a model to accurately represent a broad range of highly complex functions~\citep{raghu2017expressive}. A popular approach to modeling complicated nonlinear dynamical systems is to form a piecewise approximation by transitioning between a fixed set of linear regimes throughout the trajectory of the signal~\citep{Vyas2020}. 
These models include SLDS, rSLDS and their variants. A direct result of the fixed number of linear regimes is that it has limited capacity in the types of systems that it can well approximate. If more capacity is required, then we must specify a larger number of linear states, and fit the entire model from scratch. 

For example, switched systems often learn approximate dynamics which have discontinuities along the boundaries of the switches. This can be well suited to capturing sudden changes in dynamics, but may not accurately represent smooth transitions between dynamical modes. 
Approximating smooth transitions requires increasing the number of linear states available to capture intermediate stages. Additionally, learning discrete states independently becomes prohibitively costly when estimating dynamics for a population that contains multiple independent subgroups. In this scenario, SLDS learns distinct systems for each potential combination of dynamics. However, the number of possibilities grows exponentially as the number of subgroups and dynamical modes increases, rendering it combinatorially impractical. Moreover, this challenge is compounded during inference where the learned systems are rigid and cannot adapt to capture similar dynamics. Instead, SLDS must learn new distinct states to accurately represent subtle variations of similar dynamics. For a fixed number of parameters, dLDS improves the expressivity of the switching linear approach by offering a controlled way to flexibly modify and reuse learned linear regimes. 

\begin{figure}[t]
    \centering
    \includegraphics[width=0.95\textwidth]{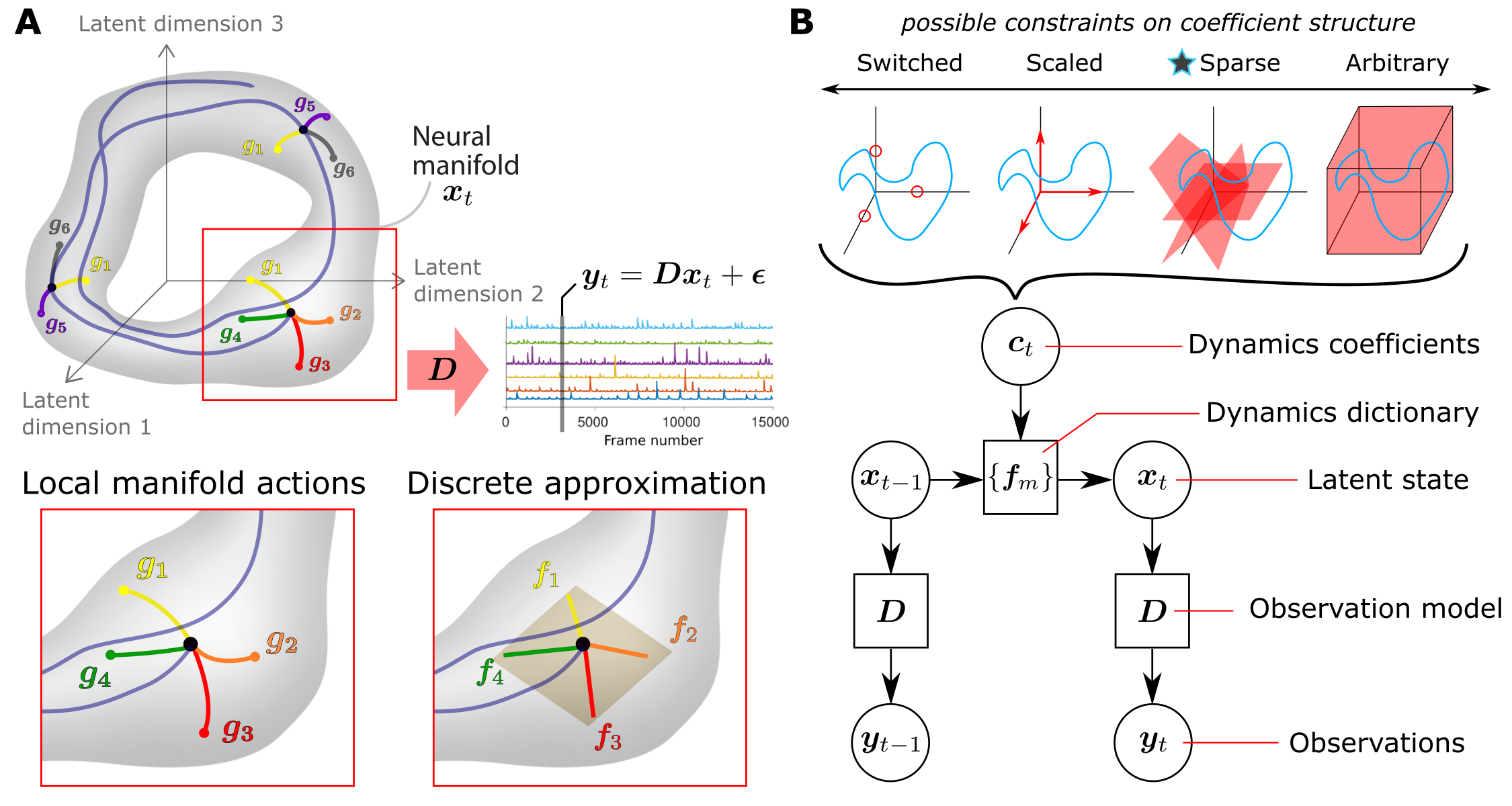}
    \caption{\textbf{Decomposed dynamical system model.} \textbf{A:} Trajectories along the manifold are guided by local DOs.
    In neuroscience, we indirectly observe the latent manifold state $\bm{x}_t$ through the observation model $\bm{D}$. The space of transports $\{\bm{g}_l\}_{l=1:L}$ can be learned directly or through a discretized approximation $\{\bm{f}_m\}_{m=1:M}$. \textbf{B:} The decomposed linear dynamical systems includes an observation model, a dynamics model, and hierarchical variables $\bm{c}_t$ that control the non-stationarity in the dynamics. These dynamics coefficients can be structured (top), e.g., one fixed active coefficient at a time results in switching between discrete states, whereas enabling flexibility in the coefficient's value can enable scaled dynamics, sparsely structured dynamics, or even more arbitrarily distributed dynamics.}
    \label{fig:introFig}
\end{figure}

\section{Decomposed linear dynamical systems}

We begin our model description with the observation model.   
Let $\bm{Y} = [\bm{y}_1,...,\bm{y}_T] \in \mathbb{R}^{k \times T}$ be a sequence of $T$ $k$-dimensional observation vectors $\bm{y}_1,...,\bm{y}_T$. 
A loading matrix $\bm{D} \in \mathbb{R}^{k\times p}$ links each observation vector $\bm{y}_t$ to its underlying latent state, $\bm{x}_t \in \mathbb{R}^{p}$, such that: \\
\begin{equation}  \label{eq:1}
 \bm{y}_t = \bm{D}\bm{x}_t + \bm{\epsilon}_t,
\end{equation}
where $\bm{\epsilon}_t$ is measurement noise from an isotropic Gaussian distribution.

Critical to dLDS is modeling the temporal evolution of the latent states along the underlying geometry. We begin with the common assumption that $\bm{x}_t$ lies on a $d$-dimensional manifold $\mathcal{M}\subset\mathbb{R}^p$. To stay on the manifold over time, the flows guiding the latent state $\bm{x}_t$ must move according to the local symmetries $\bm{g}_l$ of the manifold that define the manifold tangent space at each point. An appropriate continuous-time model that describes the movement along the manifold can be expressed as $\dot{\bm{x}}_{t} = \bm{G_t}\bm{x}_t$ where $\bm{G}_t$ represents a local transformation at any point in time that maps the point $\bm{x}_t$ onto nearby points on the manifold. Taking a finite step integrates a first-order linear differential equation whose solution involves the matrix exponential.
Moreover, as the tangent space is a subspace, the operator $\bm{G}_t$ can be decomposed at each time point into a linear combination of dictionary elements $\bm{g}_l$ which we refer to as dynamic operators (DOs) to emphasize that they are specifically designed to represent dynamic processes rather than static geometry. The set of DOs span the space of possible local motions at different points on the manifold (i.e., $\bm{g}_l\bm{x}_t$ spans the tangent space at $\bm{x}_t$) and are weighted by the coefficients $\bm{c}_{t} = [c_{1t},...,c_{Lt}]^T$ which encode the representation of the dynamical trajectory through $\bm{x}_t$. Any trajectory between two points $\bm{x}_{t}$ and $\bm{x}_{t+\tau}$ on the manifold, can be written as
\begin{align}
        \bm{G}_t &= \sum_{l=1}^L\bm{g}_lc_{lt} \\  
    \label{eqn:contDyn}
    \bm{x}_{t+\tau} &= \textrm{expm}\left(\bm{G}_t\tau\right)\bm{x}_t.   
\end{align}

As a result, the latent dynamics, $\bm{x}_t \in \mathbb{R}^{p}$, evolves according to the non-stationary dynamical systems matrix ($\bm{G}_t$) defined by the evolution of $\bm{c}_t$ over time. The continuous-time differential formulation in Equation~\eqref{eqn:contDyn} can hence flexibly define dynamics. However, it requires computing a matrix exponential, which can be computationally expensive in higher dimensions. Thus, we can approximate the action of the DO dictionary over a local area as $\exp(\sum_{l=1}^L\bm{g}_lc_{lt}) \approx \bm{F}_t$ for each time point $t = 1, ... T$, which can be expanded into its own basis as 
\begin{equation} \label{eq:3}
 \bm{F}_t = \sum_{m=1}^M \bm{f}_m c_{mt} + \nu_t,
\end{equation}
where $f_m \in \mathbb{R}^{p \times p}$ for all $m = 1 \dots M$ the dynamics coefficients for the discrete time form are denoted by $\bm{c}_{t} = [c_{1t},...,c_{Mt}]^T$, and the dictionary is composed of a set of $M$ linear dynamical systems represented by$\{\bm{f}_m\}_{m=1:M}$.
Note that we use here $\bm{c}_{t}$ as the dynamics decomposition coefficients to maintain consistency when discussing the movement along the manifold in the latent space. The discrete-time dynamics model is defined as 
\begin{equation}  \label{eq:discrete_examples}
\bm{x}_{t} = \bm{F}_t\bm{x}_{t-1} = \sum_{m=1}^M \bm{f}_m c_{mt}\bm{x}_{t-1}. 
\end{equation}

In dLDS, we distinguish between the two sets of unknowns: model parameters ($\bm{D}$ and either $\{\bm{g}_l\}_{l=1:L}$ or $\{\bm{f}_m\}_{m=1:M}$) and model coefficients ($\bm{x}_t$ and $\bm{c}_t$). The model parameters define the total geometry of the data that we wish to learn through example data. The model coefficients dictate specific trajectories through the learned geometry that can be inferred given the model parameters for any data coming from the same distribution. Next, we focus on defining algorithms capable of learning the model parameters from data.

\textbf{Model training framework.} 
We frame learning the model parameters from example data as a maximum likelihood (ML) estimation:
\begin{gather}
 \widehat{\theta} = \arg\max_{\widehat{\theta}} \left[ \prod_{t=1}^T P\left(\bm{y}_t|\bm{D},  \{\bm{f}_m \}_{m=1:M}\right) \right], 
\end{gather}
where for simplicity we focus here on the discrete dynamics $\{\bm{f}_m\}_{m=1:M}$. A similar learning algorithm can be derived for continuous dynamics by including the matrix exponential into the transition, at the cost of introducing nonconvexity to the objective function. To deal with this challenge, we leverage advances in automatic differentiation and stochastic optimization to provide estimates of the coefficients and model parameters. 
As the total likelihood is not directly known, we instead consider the marginal probability that is defined by the conditional likelihoods and priors defined by the geometry-driven dynamics model
\begin{gather}
 \widehat{\theta} = \arg\max_{\theta} \int \int \left[ \prod_{t=1}^T P(\bm{y}_t|\bm{D}, \bm{x}_t) \right] \left[  \prod_{t=2}^T P(\bm{x}_t|\{\bm{f}_t\}_{i=1:M}, \bm{x}_{t-1}, \bm{c}_{t})P(\bm{c}_t|\bm{c}_{t-1}) \right] \dif \bm{x} \dif \bm{c}.
\end{gather}

In general, these integrals are intractable, rendering direct maximum likelihood estimation infeasible. Instead, we follow the DL literature~\citep{OLS:1996} which can be derived via a variational approach where the posterior over the variables we wish to marginalize over is approximated with a point-mass distribution over the posterior mode~\citep{barello2018sparse}. This approximation results in an expectation-maximization (EM) procedure where we iterate over inferring the model coefficients given an estimate of the model parameters, 
\begin{equation} \label{eq:4}
\{\bm{x}_t, \bm{c}_t\} = \arg\min_{\bm{x}_t, \bm{c}_t} -\sum_t\log(P(\bm{y}_t| \bm{D}, \bm{x}_t)P(\bm{x}_t,\bm{c}_t | \{\bm{f}_m\}, \bm{x}_{t-1},\bm{c}_{t-1})),
\end{equation}
and updating the model parameters given the estimated coefficients:
\begin{eqnarray}
\bm{D} & \leftarrow & \Pi_{C_D}{(\bm{D} + \eta_D\nabla_{\bm{D}} \sum_t\log(P(\bm{y}_t| \bm{D}, \bm{x}_t)P(\bm{x}_t,\bm{c}_t | \{\bm{f}_m\}, \bm{x}_{t-1},\bm{c}_{t-1})))} \\
\bm{f}_m & \leftarrow & \Pi_{C_{f_m}}{(\bm{f}_m + \eta_f\nabla_{\bm{f}_m}\sum_t \log(P(\bm{y}_t| \bm{D}, \bm{x}_t)P(\bm{x}_t,\bm{c}_t | \{\bm{f}_m\}, \bm{x}_{t-1},\bm{c}_{t-1})))},
\end{eqnarray}
where $\eta_D$ and $\eta_f$ are the gradient-descent step sizes that need not be the same for $\bm{D}$ and $\bm{f}$, nor be constant over the learning process, and $\Pi_{C_D}$ and $\Pi_{C_{f_m}}$ are the projection operators to the constraint sets $C_D$ or $C_{f_m}$ on $\bm{D}$ or $\bm{f}_m$ respectively. These constraint sets can serve as another sparsity constraint, refer to matrix normalization, or just be the identity operator for the unconstrained case. We leave the constraint set generic as it may depend on a given specific application's goal. 

\textbf{Expectation step: estimating the model coefficients.} The first step in running the EM approach is to derive an efficient process for inferring $
\bm{x}_t$ and $\bm{c}_t$ given the data and an estimate of the model parameters. Assuming isotropic Gaussian noise in the observation and dynamics models, the inference step for the discrete time setting can be written as
\begin{gather} \label{eq:5}
\{\widehat{\bm{x}}_{t}, \widehat{\bm{c}}_t\}_{t=1}^T = \arg\min_{\{\bm{x}_t, \bm{c}_t\}} \left[\sum_{t=1}^T  \left\| \bm{y_t} - \bm{D}\bm{x}_{t}\right\|_2^2  
+\sum_{t=2}^T\lambda_0\left\| \bm{x}_{t} -\sum_{m = 1}^M {\bm{f}_m c_{mt}} \bm{x}_{t-1} \right\|_2^2 \right. \nonumber \\
\qquad\qquad\qquad\qquad\qquad\qquad\qquad\qquad\qquad\left.+ \sum_{t=1}^T\left(\lambda_1 \|\bm{x}_t\|_1  + \lambda_2 \|\bm{c}_t\|_1\right) +  \sum_{t=2}^T\lambda_3 \|\bm{c}_t - \bm{c}_{t-1}\|_2^2 \right], 
\end{gather}
where $\lambda_0$,  $\lambda_1$, $\lambda_2$, and $\lambda_3$ are regularization terms that penalize the deviations from the dynamical systems model, the sparsity of $\bm{x}_t$, the sparsity of $\bm{c}_t$, and the temporal-smoothness of the coefficients $c_t$, respectively. The $\ell_1$ sparsity regularization encourages the use of only a few dynamical systems, $\bm{f}_{m}$, at each time point. Similarly, in cases where we believe that the latent states are independent, $\ell_1$ regularization over $\bm{x}_t$ encourages appropriate decoupling. Modifying $\lambda_0$,  $\lambda_1$, and $\lambda_2$ (or setting them to zero) allows our approach to adapt to different modeling conditions by modulating the expected sparsity in both the dynamics coefficients and latent state (Fig.~\ref{fig:introFig}B, top).
The last term, $\lambda_3 \|\bm{c}_t - \bm{c}_{t-1}\|_2^2$, encourages smoothness over the time-varying coefficients. This smoothness constraint between consecutive time points can prevent high jumps and abrupt changes in the model coefficients over time that may be due to noise, thereby enhancing its ability to capture underlying trends without being overly influenced by noise or outliers. 

Solving the above for all time points can be computationally intensive, in particular, due to the bilinear form induced by the products between $\bm{x}_{t-1}$ and $\bm{c}_t$. We adopt the recent Basis Pursuit De-Noising with Dynamical Filtering (BPDN-DF)~\citep{charles2016rwl1df} approach for dynamic filtering of sparse signals. BPDN-DF is used for inferring the coefficients, $\bm{c}_t$, and the latent variables, $\bm{x}_t$, at a single time point, and proceeds through all time points sequentially. By conditioning on past time point estimates, the dynamics prediction $\sum_{m = 1}^M {c_{mt}\bm{f}_m} \widehat{\bm{x}}_{t-1}$ can be written as $\widetilde{\bm{F_t}}\bm{c}_{t}$ where the $m^{th}$ column of $\widetilde{\bm{F}}\in\mathbb{R}^{p\times M}$ is the product $\bm{f}_m\widehat{\bm{x}}_{t-1}$. Thus, the dynamics can be then rewritten as 
\begin{equation} \label{eq:xtplus1}
    \bm{x}_{t} = \bm{F}_t\bm{x}_{t-1} = \sum_{m=1}^M c_{mt} \bm{f}_m \bm{x}_{t-1} = \sum_{m=1}^M [\bm{f}_m\bm{x}_{t-1}] c_{mt} = \widetilde{\bm{F_t}}\bm{c}_t.
\end{equation}
Inferring all coefficients at each time step reduces to a LASSO  problem (or partial LASSO or least-squares, depending on if any $\lambda$ values are set to zero), 
\begin{equation} \label{eqn:BPDNDF}
\ \widehat{\bm{x}}_t, \widehat{\bm{c}}_t = \arg\min_{\bm{x}_t, \bm{c}_t}  \left[ \left\| \bm{y_t} - \bm{D}\bm{x}_{t}\right\|_2^2 + \lambda_0\left\| \bm{x}_{t} - \widetilde{\bm{F}}_t \bm{c}_{t}\right\|_2^2  + \lambda_1 \|\bm{x}_t\|_1  + \lambda_2 \|\bm{c}_t\|_1 + \lambda_3\|\bm{c}_t - \widehat{\bm{c}}_t\|_2^2 \right].
\end{equation}

We note that for the continuous case, the same basic framework holds. However, the matrix exponential prevents the consolidation of the previous estimate and dynamics into a single matrix $\widetilde{\bm{F}}$, and retains the following form:
\begin{gather} 
\widehat{\bm{x}}_t, \widehat{\bm{c}}_t = \arg\min_{\bm{x}_t, \bm{c}_t}  \Biggl[ \left\| \bm{y} - \bm{D}\bm{x}_{t}\right\|_2^2 + \lambda_0\left\| \bm{x}_{t} - \textrm{expm}\left(\sum_{l=1}^L{\bm{g}_lc_{lt}}\right)\widehat{\bm{x}}_{t-1}\right\|_2^2 . \nonumber \\
\qquad\qquad\qquad\qquad\qquad\qquad\qquad\qquad\qquad\qquad+ \lambda_1 \|\bm{x}_t\|_1  + \lambda_2 \|\bm{c}_t\|_1 + \lambda_3\|\bm{c}_t - \widehat{\bm{c}}_t\|_2^2 \Biggr]. \label{eqn:BPDNDFcont}
\end{gather}

\textbf{Maximization step: updating the model parameters.} Given the inferred model coefficients, the second step of the learning framework is a gradient step over the model parameters. For the latent state projection $\bm{D}$, the gradient can be computed as in traditional DL~\citep{OLS:1996}: 
\begin{gather}
     \widehat{\bm{D}} \leftarrow  \Pi_{C_D}(\bm{D} - \eta_D\nabla_{\bm{D}}\sum_{t=1}^T(\bm{y}_t - \bm{D}\bm{x}_t)^2) =  \Pi_{C_D}(\bm{D} + \eta_D\sum_{t=1}^T (\bm{y}_t - \bm{D}\bm{x}_t)\bm{x}_t^T)). \label{eqn:Dupdate}
\end{gather}
As with any DL procedure, there is a fundamental ambiguity in the scale of the model parameters and coefficients. We therefore constrain each column of $\bm{D}$ to have unit-norm via the projection $\prod_{C_D}$.
Note that this update rule is appropriate in both the continuous and discrete time cases. 

For the dynamics we start with the discrete case. For each $\bm{f}_m$, the gradient can be computed as
\begin{eqnarray} \label{eqn:Fupdate}
    \widehat{\bm{f}}_m &\leftarrow & \Pi_{C_{f_m}}{\left(\widehat{\bm{f}}_m  - \eta_f \nabla_{f_m} \sum_{t=2}^T\left\|\bm{x}_t -\sum_{m = 1}^{M} \bm{c}_{mt} \bm{f}_m\bm{x}_{t-1}\right\|_2^2 \right)} \\
&=& \Pi_{C_{f_m}}{\left(\widehat{\bm{f}}_m  + \eta_f\sum_{t=2}^T \left(\bm{c}_{mt}\left(\bm{x}_t -\widetilde{\bm{F}}_t\bm{c}_{t}\right)\bm{x}_{t-1}^T \right)\right)}.
\end{eqnarray} 

A similar ambiguity with the latent projection $\bm{D}$ occurs between the scale of $\bm{f}_m$ and $\bm{c}_t$, so we impose the constraint that each $\bm{f}_m$ has a unit spectral radius. This constraint is applied at each optimization iteration and in practice transforms the gradient descent to a projected gradient descent. 

Parameter updates for updating the continuous operators can be similarly derived over the gradient of the term involving the matrix exponential of $\sum_l\bm{g}_lc_{lt}$. Thus the derivative must be computed through the matrix exponential (see Algorithm~\ref{alg:ctmt}), and we find it easier to replace the strict constraint over the spectral radius with a looser regularization over the Frobenius norm $\|\bm{g}_l\|_F^2$. 
\begin{equation} \label{eq:ct.loss}
\widehat{\bm{g}}_l \leftarrow  \Pi_{C_{g_l}}{\left(\widehat{\bm{g}}_l  - \eta_g \nabla_{g_l} \sum_{t=2}^T\left\|\bm{x}_t - \textrm{expm}\left(\sum_{l=1}^L c_{lt}\bm{g}_l \right) \bm{x}_{t-1}\right\|_2^2 +  \lambda_g\sum_{l=1}^L\|\bm{g}_l\|_F^2 \right)}, 
\end{equation} 
where $\lambda_g$ is the weight of the Frobenius norm regularization over the dynamics matrices $\bm{g}_l$, and $\Pi_{C_{g_l}}$ is the projection operator to the constraint set $C_{g_l}$ on $\bm{g_l}$. 

\begin{algorithm}[t!]
\caption{dLDS Model Training}\label{alg:cap}
\begin{algorithmic}
\State \textbf{Input} $\bm{Y}$, $\lambda_1$, $\lambda_2$, $\eta$, $M$ \Comment{Input observations and hyperparameters}
\State \textbf{Initialize}  $\{\bm{f}_m\}_{m=1:M}$, $\bm{D}$ \Comment{Randomly initialize model parameters} %
\While{not converged} \Comment{Iterating until convergence}
        \State Infer $\bm{c}_t$ and $\bm{x}_t$ via Equation~\ref{eqn:BPDNDF}
        \State Update $\bm{D}$ via Equation~\ref{eqn:Dupdate} 
        \State Update each $\bm{f}_m$ via Equation~\ref{eqn:Fupdate} 
    \If{$rMSE$ does not change} \Comment{Check if stuck in a local minimum} 
            \State $\bm{f}_{m,ij} \gets \bm{f}_{m,ij} + \bm{\xi}_{ij}$ where $\bm{\xi}_{ij} \sim \mathcal{N}(0,\sigma^2)$ \Comment{Randomly perturb all elements of $\bm{f}_{m}$} 
    \EndIf
\EndWhile
\end{algorithmic}
\end{algorithm}

The full algorithm is presented in Algorithm~\ref{alg:cap}. As a special case that can be of interest in lower-dimensional data settings, we also consider the scenario in which the observation matrix, $\bm{D}$, is fixed to be the identity matrix, i.e., $\bm{y}_t = \bm{x}_t$. In this case we have direct observations of points on the manifold, and the model learning can be reduced in computational complexity (see Appendix~\ref{sec:NoObs} for the continuous and discrete cases). In all cases, we also intermittently perturb the model parameters to prevent local minima. 

\subsection{Interpretability for dLDS} 

Each individual component of the sparse decomposition is interpretable in the context of dynamical systems. For the continuous coefficients, the scale and magnitude have intuitive meaning and can modify the dictionary to account for many of the natural variations seen in time series data. Changes in speed or frequency of oscillations can be represented by increasing the coefficient’s magnitude. Larger coefficients represent faster speeds and higher frequencies while smaller magnitudes represent slow speeds or low frequency oscillations. The signs of the coefficients also represent the direction of motion in the dynamical systems. A positive sign indicates movement in the direction of the learned system while a negative sign indicates reverse-time or backwards movement.  

The dictionary is also readily interpreted since linear dynamical systems have analytical solutions which fully characterize its properties mathematically. The rank and structure of the linear systems can reveal the underlying dimensionality of a particular dynamic. A system with low rank and block-diagonal structure may suggest subspace-specific transitions and enables the modeling of trajectories generated from several independent subnetworks. In addition to the interpretability of individual model components, it is also possible to reason over the model as a whole. By enforcing the sparsity constraint over the coefficients, we encourage the model to reduce redundancy and learn dictionary elements that are statistically independent.  Moreover, sparsity reduces the complexity of the representation and facilitates a clearer understanding of the underlying dynamics by allowing only a few active components at any point in time. Furthermore, we formulate a mixture model to represent dynamics, and enable the learning of shared dynamics, where multiple processes are governed by the same underlying system.  In grouping together similar processes and modeling them as a single unit, we simplify the analysis and interpretation of the data. 

\subsection{Expressivity for dLDS}

dLDS improves the expressiveness of traditional switching linear models by replacing the discrete dynamical states with a sparse coding model. By relaxing the discrete switches and introducing continuous coefficients, we enable the model to compactly represent a more flexible family of linear approximations for nonlinear dynamical systems during learning and inference, as compared to switch-based models. In fact there are several types of constraints we can impose on the coefficients (Fig.~\ref{fig:introFig}B). In the most restrictive setting, coefficients are constrained to being 1-hot vectors, which recovers the SLDS model as a special case. Relaxing the constraint on the magnitude of the coefficients results in a model with scaled coefficients over individual systems (Fig~\ref{fig:introFig}B). As a result of the 1-sparse coefficients, only a single system can be active, but its magnitude can be scaled positively and negatively. Time-warped AR-HMM~\citep{costacurta2022distinguishing} is an example of a model in this family and uses a single continuous coefficient to scale existing discrete states to compactly represent learned dynamics at different speeds. dLDS relaxes the constraint of only having a single system active at a time and allows for n-sparse representations of the linear approximation at each point in time. Allowing multiple dynamics components to be active at a time dramatically improves the expressivity of dLDS in a way that is similar to how Factorial HMMs are able to represent an exponential number of bits given a linear increase in the number of parameters~\citep{ghahramani1995factorial}. 

For a fixed number of parameters, a more expressive model should be able to achieve lower approximation error over a broad range of systems~\citep{dong2020expressivity}. During inference, dLDS achieves lower approximation error through its continuous sparse coefficients, which can flexibly adjust to modify and combine existing learned dynamics to accurately capture new trajectories from similar systems. By contrast, switching linear systems cannot adapt learned discrete states to unseen trajectories even in the presence of small variations. During learning, dLDS offers a rich model that can efficiently capture important dynamics such as smooth transitions between states and ramping of amplitudes. The SLDS model, on the other hand, can achieve a similar expressivity but only in the limit of having a separate discrete state for every pair of consecutive time slices. Such a representation is prohibitively expensive, scaling linearly with the length of the signal and quadratically with the number of discrete states. Moreover, increasing the number of parameters in this manner would result in severe overfitting and prevent SLDS models from effectively capturing interesting structures and learning meaningful representations. Further, allowing for a large number of discrete states introduces discontinuities on the boundaries of the linear dynamics which can be an inappropriate assumption for signals collected, e.g., from neural systems.  

\section{Experiments}
To demonstrate the capabilities of dLDS, we present several experiments in both continuous and discrete settings. We first showcase the model's efficiency in representing continuous data (Sections~\ref{subsec:ContinuousSpeed},~\ref{subsec:ContinuousEfficTraj}, Fig.~\ref{fig:contDyn}). Then we demonstrate the model's ability to reconstruct discrete dynamics and recover shared underlying operators from synthetic, low-dimensional, nonlinear dynamical systems in the case of changing stability regimes (Section~\ref{subsec:discretestability}, Fig.~\ref{fig:2_spirals}), smooth transitions between ground truth DOs (Section~\ref{subsec:discretesmooth}, Fig.~\ref{fig:2_spirals}), independent but simultaneously observed systems (Section~\ref{subsec:discretesimult}, Fig.~\ref{fig:matlab}), and across model regularization settings and initializations (Sections~\ref{subsec:discreteL1},~\ref{subsec:discreteIC}, Figs.~\ref{fig:fhn},~\ref{fig:lorenz}). Finally, we apply the model to real-world \textit{C. elegans} calcium imaging data to uncover underlying patterns that were previously obscured (Section~\ref{subsec:Celegans}, Figs.~\ref{fig:SupplCE},~\ref{fig:celegans}).

\subsection{Continuous-time dLDS can efficiently model dynamics at different speeds} \label{subsec:ContinuousSpeed}
\begin{figure}[t!]
    \centering
    \includegraphics[width=0.95\textwidth]{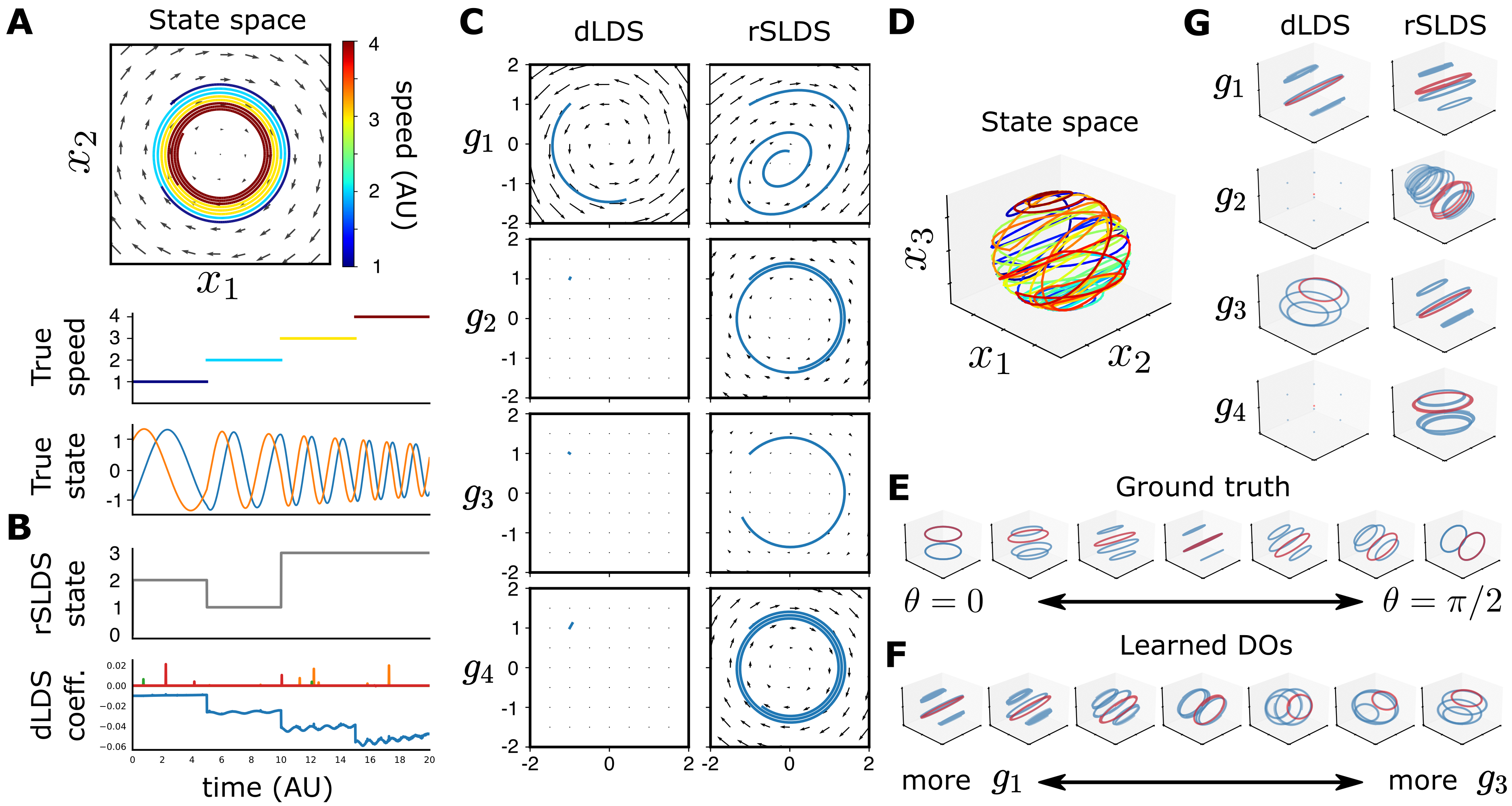}
    \caption{\textbf{Synthetic linear systems, examples of efficient representation.}
    \textbf{A:} The generated path from ground truth 2D spiral system, colored by speed in phase space (top) and unrolled over time (bottom). 
    \textbf{B:} The inferred rSLDS discrete states (top) and the inferred dLDS coefficients (bottom).
    rSLDS models speed changes with three discrete states and incorrectly groups the two fastest speeds together while dLDS changes coefficients on a single dictionary element. 
    \textbf{C:} dLDS learns a single dictionary element (left column) that can be reused while rSLDS learns redundant systems (right column).
    \textbf{Smooth transitions between DOs represent different paths on a spherical manifold:}
    \textbf{D:} The generated path on the ground truth 3D sphere colored to visualize progression through time.
    \textbf{E:} Possible ground truth rotations. Multiple traces show dynamics and a single trace is highlighted (red) for better visibility. 
    \textbf{F:} Convex combinations of learned DOs $g_1$ and $g_3$ allow for smooth transitions along continuum of rotated systems.
    \textbf{G:} dLDS learns two DOs that can be combined to represent all paths on the sphere while rSLDS must learn each angle of rotation separately.
    }
    \label{fig:contDyn}
\end{figure}

Continuous coefficients provide an effective means for efficient representation and smooth transitions between dynamical modes. Specifically, by modulating the dynamics coefficients $\bm{c}_t$, dLDS can accurately model the same system operating at different speeds. To demonstrate the representational efficiency, we consider a two-dimensional slowly decaying spiral that progressively increases its rotational velocity by one unit of speed every 5 units of time. The trajectory sampled from this system, as depicted in Figure~\ref{fig:contDyn}A, undergoes four speed shifts. In our study, we compare dLDS with standard switching linear models, focusing on rSLDS~\citep{linderman2017bayesian}. To ensure fair comparisons, we approximately matched the number of parameters and fit the dynamics using $L=4$ systems. dLDS learned to represent the entire system using a single DO and effectively eliminated unused operators (Fig.~\ref{fig:contDyn}C left column). Furthermore, dLDS used the coefficients to modify the single DO in order to capture the changes in speed (Fig.~\ref{fig:contDyn}B bottom). By contrast, rSLDS recovered a less efficient representation by learning four non-zero discrete states, despite only using three of them during inference (Fig.~\ref{fig:contDyn}C right column).  Although rSLDS managed to capture the original system, it was unable to identify the shared behavior and used separate states to model the same dynamics at different speeds. Lastly, it is worth noting that rSLDS encountered limitations in inference due to its rigid discrete states, which obscured when the system changes between the fastest speeds as the result of approximation errors in the dynamics (Fig.~\ref{fig:contDyn}B top). On the other hand, dLDS exhibited adaptability by flexibly adjusting its coefficients to identify speed changes, despite approximation errors in the DO.

\subsection{Continuous-time dLDS efficiently represents trajectories on a sphere}\label{subsec:ContinuousEfficTraj}

Next, we highlight the ability of dLDS to model smooth transitions in dynamics of a particle moving around an axis of a spherical manifold. At regular intervals of 5 units of time, the system undergoes a random rotation $\theta \in [0,2\pi]$ along the x-axis.
Figure~\ref{fig:contDyn}D shows a sampled path for 100 rotations while Figure~\ref{fig:contDyn}E illustrates the continuum of ground truth rotations that can occur. In contrast to the previous experiment where we fit each model with the exact number of ground truth modes, we consider the situation where the models are fit with fewer systems than the total number of dynamical modes. In both switched and decomposed models, we fit the dynamics using $L=4$ systems. 

dLDS recovered two DOs, which represented the original system at two distinct angles , and shrank the remaining operators to zero (Fig.~\ref{fig:contDyn}G, left). The two DOs offered an efficient description of the system, as they could be combined to represent the complete range of rotations (Fig.~\ref{fig:contDyn}F). By smoothly changing the ratios of the coefficients, dLDS could smoothly transition between each DO to accurately represent intermediate trajectories. By contrast, rSLDS learned four non-zero discrete states, each capturing the original system at different angles (Fig.~\ref{fig:contDyn}G, right). Although rSLDS may provide an accurate description of the particle when its trajectory aligns with one of the learned states, trajectories that fall in between one of the learned discrete states may result in poor approximation and incorrectly inferred states. Again, rSLDS learned each discrete state separately and was unable to find the relationship between them, while dLDS learned that similar trajectories along the spherical manifold can be represented by sharing DOs.

\subsection{Discrete-time dLDS flexibly models different stability regimes}\label{subsec:discretestability}

By adjusting coefficients, the discrete-time dLDS model enables modeling shifts in the stability of a system. We consider a simple transition of a spiral system that shifts from a stable, decaying regime to an unstable, expanding regime:
\begin{equation}\label{eq:discrete_spirals}
x_t = 
\begin{cases}
0.99\bm{f}\bm{x}_{t-1} & \text{if } 0 < t \leq \frac{T}{2}, \\
\frac{1}{0.99}\bm{f}\bm{x}_{t-1} & \text{if } \frac{T}{2} < t \leq T.
\end{cases}
\ \bm{f} =
\begin{bmatrix}
\cos(\theta) & \sin(\theta) \\
-\sin(\theta) & \cos(\theta)
\end{bmatrix}
\quad \text{where } \theta = \frac{\pi}{5}.
\end{equation}
In this system, the rotational matrix $\bm{f}$ underlies the ground truth dynamics, and the matrix's coefficient $c_t$ has a fixed value of $c_t = 0.99$ for $0 < t \leq T/2$ and then the inverse value $c_t = 1/0.99$ for $T/2 < t \leq T$ (Fig.~\ref{fig:2_spirals}A blue). 
The rotational nature of $\bm{f}$ generates the spiral that initially converges exponentially and then moves outward, becoming unstable (Fig.~\ref{fig:2_spirals}B blue). 

We trained both rSLDS and dLDS on a single ($M=1$) dynamical state to emphasize the efficiency in representation. In this experiment, dLDS well approximates this original system (Fig.~\ref{fig:2_spirals}B red) by using the identified coefficients (Fig.~\ref{fig:2_spirals}A red) to adjust the single learned DO's stability. In doing so, it is able to efficiently control the direction of the rotation (inwards or outwards) while retaining the system's overall rotational behavior. Conversely, fitting rSLDS to a single discrete state conceals the presence of a shift in stability since learned dynamics are rigid during inference (Fig.~\ref{fig:2_spirals}B green).

In fact, we find that dLDS can recover the ground truth dynamics operator with high precision and accurately reconstruct the dynamics to a level not achievable in a switched model with a single operator. 


\subsection{Discrete-time dLDS efficiently reconstructs smooth transitions between ground truth dynamics} \label{subsec:discretesmooth}

\begin{figure}[t]
\centering
\includegraphics[width=1\textwidth]{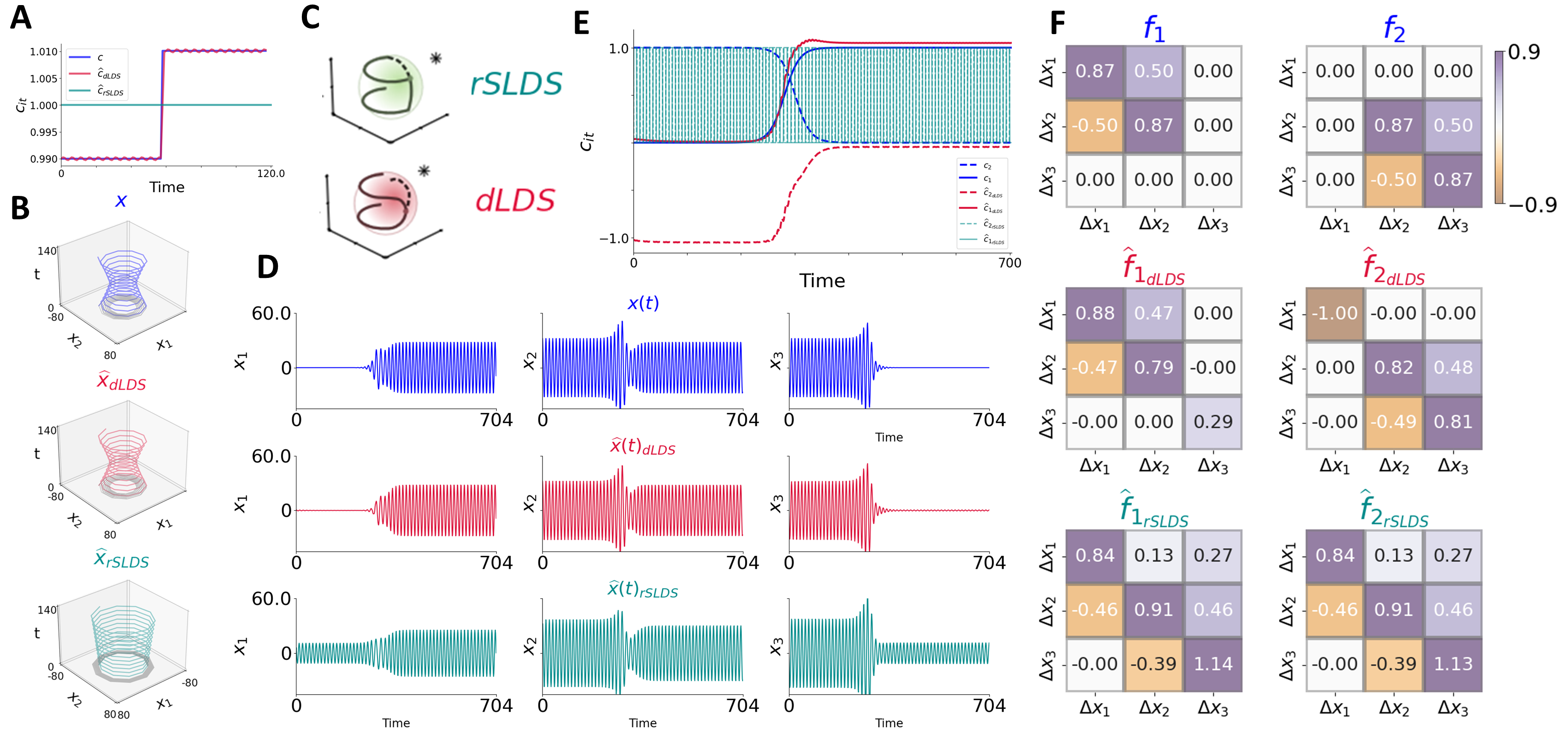}
\caption{
\textbf{A,B: dLDS captures changes in system stability.}
\textbf{A:} Comparison of the generated coefficients (blue) versus the dLDS recovered coefficients (red), and the rSLDS coefficients (green). 
\textbf{B:} The ground truth dynamics over time (blue, top), versus the recovered dynamics by dLDS (middle, red) and rSLDS (bottom, green). 
\textbf{C,D,E,F: dLDS can model smooth transitions in discrete-time dynamics.}
\textbf{C:} Schematic behavior of dLDS and switching dynamics in modeling transitions between linear dynamics. Switching systems jump between dynamic states producing sharp trajectories while dLDS can smoothly change the DO coefficients to capture gradual transitions in the system. 
\textbf{D:} The ground truth dynamics (top, blue), compared to the recovered dynamics by dLDS (middle, red) and rSLDS (right, green). Each column corresponds to a different axis.
\textbf{E:} Comparison of the generated coefficients (blue), the dLDS recovered coefficients (red), and the rSLDS coefficients (green). 
\textbf{F:} The ground-truth DOs (top, blue) versus the DOs recovered by dLDS (middle, red) and rSLDS (bottom, green). 
}
\label{fig:2_spirals}
\end{figure}

While switching models demonstrate excellent performance in capturing sudden changes in dynamics, the discrete nature of their dynamical states presents a challenge when it comes to representing seamless transitions. By contrast, discrete-time dLDS can model smooth transitions between two operational modes of a system (Fig.~\ref{fig:2_spirals}C). To illustrate this, we construct a system generated from a set of two rotational linear operators $\bm{f}_1$ and $\bm{f}_2$ (Fig.~\ref{fig:2_spirals}F top) whose coefficients $c_{1t}$ and $c_{2t}$ smoothly change over time following a sigmoid and a mirrored sigmoid (Fig.~\ref{fig:2_spirals}E blue), respectively: $\bm{x}_t = (c_{1t}\bm{f}_1 + c_{2t}\bm{f}_2) \bm{x}_{t-1}$ (Fig.~\ref{fig:2_spirals}D blue). dLDS trained with two DOs ($M=2$) is able to both fully reconstruct the dynamics (Fig.~\ref{fig:2_spirals}D, red) and recover the rotational part of the ground truth operators (Fig.~\ref{fig:2_spirals}F, middle). We observed that the differences between the lower right corner of $\bm{f}_1$ and the upper left corner of $\bm{f}_2$ in the reconstructed dLDS operators, compared to the real DOs, do not affect the reconstruction process. This is because the dynamic here ($\bm{x}$) is defined as a transition from a horizontal to a vertical spiral, resulting in $x_3$ and $x_1$ values that are close to zero when $\bm{f}_1$ and $\bm{f}_2$ are respectively active. 
 
When trained with two discrete states, rSLDS, by definition, cannot reconstruct such dynamics well (Fig.~\ref{fig:2_spirals}D, green). Indeed, the learned discrete states (Fig.~\ref{fig:2_spirals}F, bottom) did not recover the ground truth systems and instead learned two similar discrete states that both incorrectly combined the dynamical modes together. As a result, rSLDS inaccurately inferred erratic jumps between the two similar systems in an attempt to capture the smooth transition (Fig.~\ref{fig:2_spirals}E, green). In general, it is unclear how to appropriately set the number of discrete states to get an accurate representation of smooth transitions for switching models. Even if given an adequate number of discrete states, the intermediate stages can experience overfitting as a result of the low number of samples in the transition interval. dLDS addresses these issues through the continuous coefficients, which flexibly modify existing dynamics to account for minor variations and enable learning of shared structure even in the intermediate stages of a smooth transition. 

\subsection{Discrete-time dLDS disentangles simultaneously observed systems}\label{subsec:discretesimult}

One benefit of decomposing dynamics is the ability to account for multiplexed sub-systems within the same recorded data. Consider a single recording of a neural population that consists of two distinct sub-populations, denoted as $a$ and $b$. The sub-population $a$ consists of $N_1$ neurons, while the sub-population $b$ consists of $N_2$ neurons, and the recordings of these sub-populations over time are represented by $\bm{x}_t^a \in \mathbb{R}^{N_1 \times T}$ and $\bm{x}_t^b \in \mathbb{R}^{N_2 \times T}$, respectively. To represent the full data, we concatenate these two populations' recordings vertically, resulting in $\bm{x}_t = [\bm{x}_t^a; \bm{x}_t^b] \in \mathbb{R}^{(N_1 + N_2) \times T}$. If the switches of both sub-populations are not synchronized, even when both locally adhere to a switched LDS model, an SLDS model would need to transition whenever either subsystem $a$ \emph{or} subsystem $b$ switches. Consequently, the SLDS model would have to consider all possible coexistence scenarios of these two systems collectively. Thus, the timescales of the unique populations are lost, and the interpretability of the combined recordings becomes muddled. dLDS, however, can naturally account for such settings by summing operators that guide only the dynamics of subsets of the system. 

\begin{figure}[t]
\centering
\includegraphics[width=0.95\textwidth]{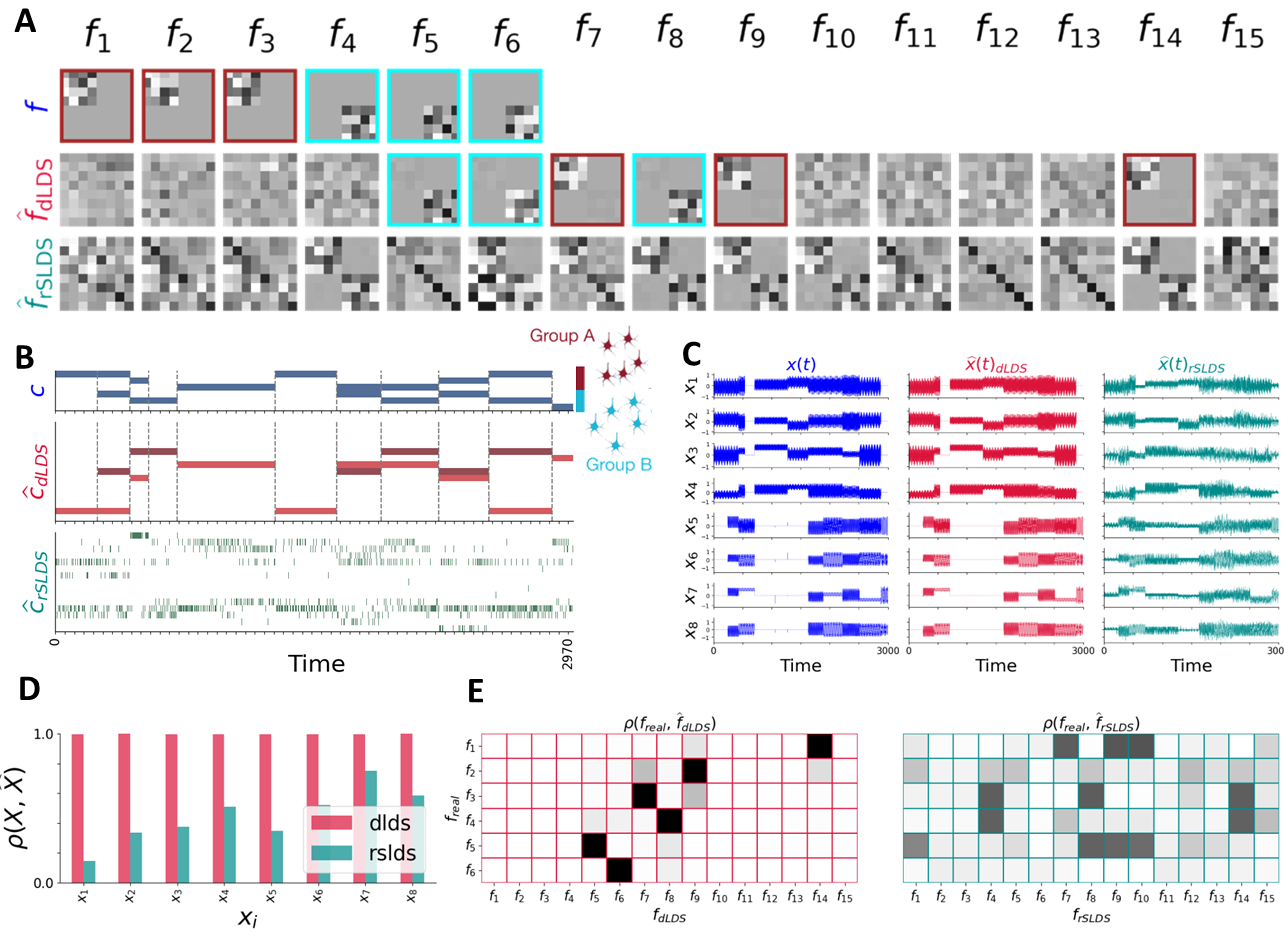}
\caption{\textbf{dLDS identifies independently evolving groups from combined time series.}
\textbf{A:} The ground truth DOs displaying the block structure of the data (top), DOs recovered by dLDS (middle), DOs recovered by rSLDS (bottom). Framed dLDS DOs have a perfect correlation to the ``true'' DOs for both populations (cyan, maroon).
\textbf{B:} The ground truth coefficients (top), dLDS recovered coefficients (middle) and rSLDS recovered coefficients (bottom). dLDS can accurately recover the structure of the ground truth and nullifies redundant coefficients, while rSLDS combines dynamics across the blocks. 
\textbf{C:} True generated dynamics (left), dLDS reconstruction (middle), and rSLDS reconstruction (right). 
\textbf{D:} The data reconstruction correlations for dLDS (red) and rSLDS (green) with the ground truth. dLDS achieves perfect reconstruction. 
\textbf{E:} The correlations of each true DO (rows) with each recovered DO (columns). dLDS (left) recovers all true DOs (each row presents exactly one black cell), while rSLDS tends to combine DOs across the two independent groups. 
}
    \label{fig:matlab}
\end{figure}

To demonstrate this effect, we simulate a ten-dimensional state $\bm{x}_t$, where the first five elements of the vector constitute ``population $a$'' and the last five elements ``population $b$''. We then generate six ground truth DOs (Fig~\ref{fig:matlab}A, top row), three of which only act on population $a$ (maroon), and the other three on population $b$ (cyan). To simplify the setting, each population constituted an SLDS system and switched abruptly between one of its three systems, or went silent with no dynamics active (Fig.~\ref{fig:matlab}C blue). We repeated this process, generating 50 draws of the process with different initial states and different switch patterns. 

We fit both dLDS and rSLDS to the generated data, providing both with a maximum of 15 dynamical operators ($M = 15$). dLDS was able to recover a basis of meaningful operators (Fig.~\ref{fig:matlab}A, middle row). Specifically, the recovered dynamical operators were well localized in terms of being localized to either the upper or lower diagonal block, indicating that dLDS learned the underlying block structure of the system. Furthermore, the dynamical operators for each learned system matched one of the ground truth operators with a correlation of $\approx 1$ (Fig.~\ref{fig:matlab}E, left), and the coefficients directly convey when each subsystem switched independent of the other subsystem (Fig.~\ref{fig:matlab}B, red). By contrast, rSLDS learned dynamical systems that had both upper and lower blocks on the diagonal active, indicating that it was learning combined dynamics across both subsystems (Fig.~\ref{fig:matlab}A, bottom row) that do not convey the true underlying independence between the subsystems. In fact, when provided with more discrete states than necessary, rSLDS learns several redundant dynamics as a result of its inability to share information between its states. For instance, $\bm{f}_4$ and $\bm{f}_{14}$ resemble each other as well as $\bm{f}_9$ and $\bm{f}_{10}$, and $\bm{f}_5$ $\bm{f}_{11}$ and $\bm{f}_{12}$ (Fig.~\ref{fig:matlab}E, right). These redundant dynamics cause unstable inference of the discrete states and result in inappropriate jumps between states with similar dynamics in the middle of the true switching intervals (Fig.~\ref{fig:matlab}B, green). 

\subsection{Discrete-time dLDS efficiently represents a range of simulated behaviors and aligns DO coefficients to the axes for interpretability}\label{subsec:discreteL1}

\begin{figure}[t!]
    \centering
    \includegraphics[width=1\textwidth]{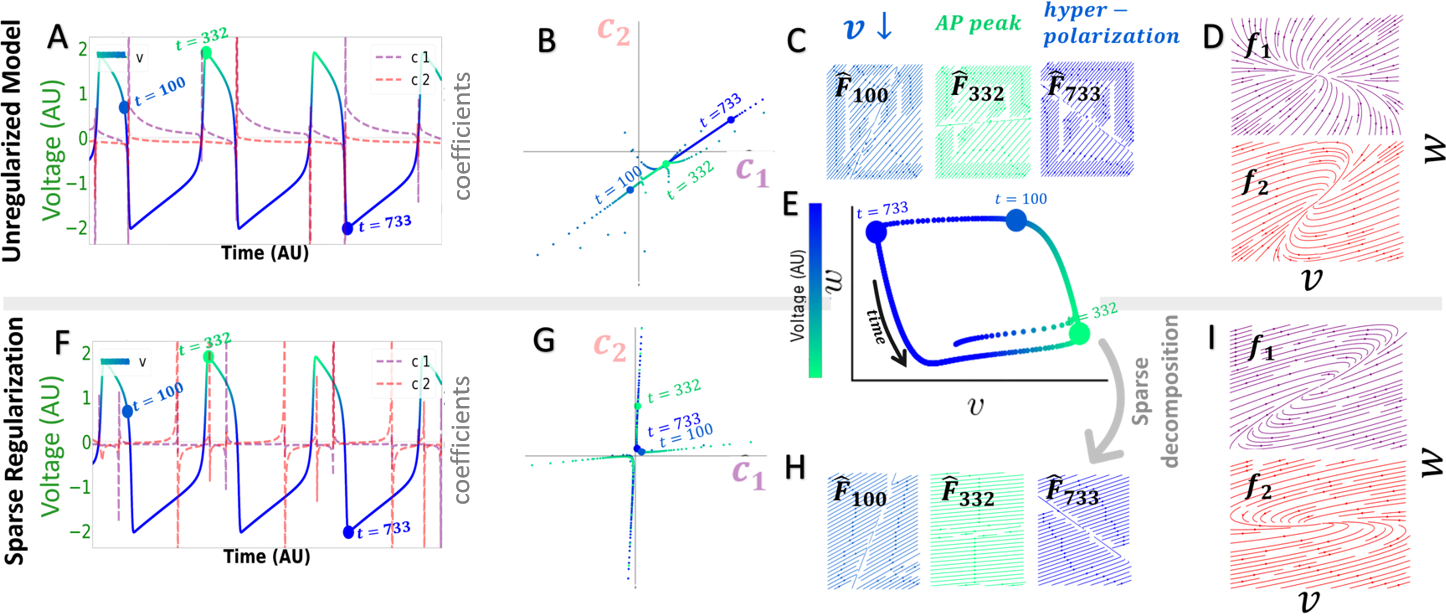}
    \caption{\textbf{Comparison between the unregularized dLDS (top row) and the sparse dLDS model (bottom row) for the FitzHugh-Nagumo oscillator.}
    \textbf{A \& F:} The temporal evolution of the membrane voltage (blue-green) and of the dynamics coefficients. Three points of interest were highlighted to showcase our model's capability to capture various behaviors with only 2 basis components. These points of interest include 1) the action potential (AP) repolarization ($t=100$), 2) the AP peak ($t=332$), and 3) hyper-polarization ($t=733$). 
    \textbf{B \& G:} Comparison between the coefficients' space of the unregularized and the regularized case. For the unregularized dLDS model, the model coefficients ($\bm{c}_t$) can occupy any location in space, and need not be on the axes. By contrast, when adding regularization to the model, sparse coefficients lie near the axes.
    \textbf{C \& H:} The learned reconstructed dynamics $\bm{F}_t$ at the time points of interest, highlighting dLDS's ability to infer more distinct phases than the number of sub-dynamics ($\bm{f_i}$)\textemdash a capability that is not available to linear or switching models.
    \textbf{D \& I:} Stream-plots of the basis operators learned by dLDS ($\bm{f}_1$ and $\bm{f}_2$).
    \textbf{E:} The phase-space plot of the FHN model (v-w space), with time points of interest highlighted.}
    \label{fig:fhn}
\end{figure}

We will now illustrate how dLDS can effectively capture intuitive aspects of nonlinear models of the underlying dynamics of biological systems. Specifically, we consider the FitzHugh–Nagumo (FHN) model: a 2D excitable-oscillatory dynamical system model that describes the temporal behavior of a nerve membrane potential in response to a stimulus~\citep{fitzhugh1961impulses} and is a simplification of the Hodgkin-Huxley model~\citep{hodgkin1952quantitative}. The FHN model is defined by a pair of conjugated differential equations denoted by
\begin{equation} \label{eq:FHN}
{\frac{\partial v}{\partial t} = v - \frac{v^3}{3} - w + I_{ext} \qquad \qquad \tau\frac{\partial w}{\partial t} = v + a - bw},
\end{equation}
where the state-space $(v,w)$ are the nerve membrane potential ($v$) and the voltage recovery variable ($w$), respectively. $I_{ext}$ is an external stimulus, and $\tau$, $a$ and $b$ are model parameters. Here we set $I_{\textrm{ext}}=0.5$, $\tau=20$, $a=0.8$, and $b=0.7$, and consider a sample initialized at $(v_0, w_0) = (-0.5, 0)$. This example demonstrates the effect of sparsity regularization on the model's representation of the dynamics. We ran dLDS with and without the $\ell_1$ regularization over $\bm{c}_t$ to test how sparsity changes the interpretability of the system (Fig.~\ref{fig:fhn}) and show that including the $\ell_1$ regularization term on the coefficients promoted their orientation towards the axes (Fig.~\ref{fig:fhn}B,G). For both the regularized and unregularized cases, the transitions of coefficients between the quadrants of the coefficients' space enable our model to well represent a diverse set of voltage behaviors (e.g., depolarization, repolarization, etc.) with only $M=2$ operators (Fig.~\ref{fig:fhn}D,I). The smooth transitions using the same two operators in both forward and reverse directions cannot be captured by switched systems. This result is consistent with the $\bm{c}_t$-plane visualization in the `arbitrary' versus `sparse' cases (e.g., Fig.~\ref{fig:introFig}B).

\subsection{Discrete-time dLDS recovers shared dynamics between different simulation initializations} \label{subsec:discreteIC}

\begin{figure}[t!]
\centering
\includegraphics[width=1\textwidth]{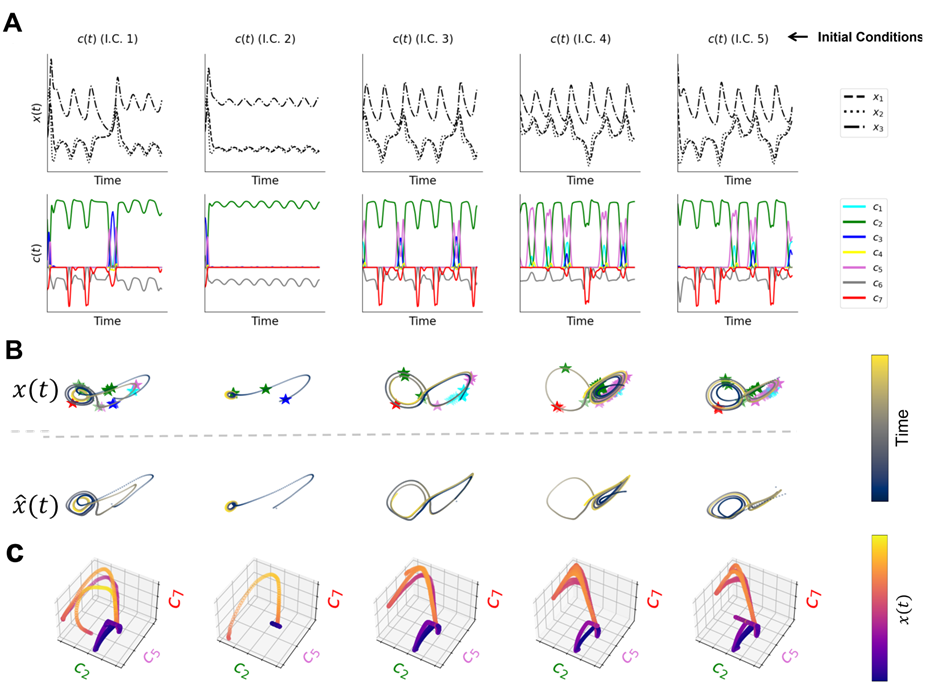}
\caption{\textbf{Demonstrating model generalizability by applying dLDS to the Lorenz attractor with unseen initial conditions (I.C.s).}
\textbf{A:} DO coefficients (bottom) plotted against the Lorenz state values (top) show coefficient patterns associated with different sections of the Lorenz attractor.
\textbf{B:} Comparison between unseen Lorenz trajectories (top row) and their dLDS reconstructions (bottom row). Stars indicate the locations of peaks of different operators' coefficients, highlighting the how underlying components localize on the attractor. 
\textbf{C:} The three most active dynamics coefficients are color-coded by the corresponding $x_1$ values. Regions visibly associated with different ranges of $x_1$ highlight repeated patterns dynamics in the Lorenz attractor.}
\label{fig:lorenz}
\end{figure}

We next demonstrate the ability of dLDS to capture the complex dynamics of the Lorenz attractor, a well-known system that exhibits chaotic behavior and is often used as a benchmark for dynamical models.
The Lorenz attractor is a nonlinear chaotic system governed by
\begin{equation} \label{eq:LORENZ}
{\frac{\partial x}{\partial t} = \alpha (y-x) \qquad \qquad \frac{\partial y}{\partial t} = x(\beta - z) - y \qquad \qquad \frac{\partial z}{\partial t} = xy - \gamma z }.
\end{equation}
For our dLDS experiment, we set $\alpha = 10$; $\beta = 25$; $\gamma = 2.67$. Here, we demonstrate dLDS' inference capability to model unseen data of the same dynamics with different initializations, using the learned set of operators. By training dLDS on a set of Lorenz attractors with different initial conditions, and then benchmarking on unseen attractors that originated from unused initial conditions, we highlight the ability of operators learned via dLDS to generalize past individual trajectories.
dLDS was able to reconstruct the unseen Lorenz attractors with high accuracy (Fig.~\ref{fig:lorenz}B bottom row). Although the model was allowed to use up to seven operators, no more than 2-3 operators were active at a time (Fig.~\ref{fig:lorenz}A bottom row), maintaining the balance between interpretability and expressivity. Additionally, the clear association between the patterns of the DOs' coefficients and the location on the attractor highlights the model's interpretability, making it a powerful tool for modeling nonlinear dynamical systems. 
In Figure~\ref{fig:lorenz}B, the stars on the Lorenz attractors mark the peaks of $c_1$ (cyan), $c_2$ (green), $c_5$ (pink), $c_7$ (red), highlighting their repeated and consistent locations along the Lorenz manifold. 
The link between the different dimensions of the Lorenz and the $\bm{c}$ values can be easily seen (Fig.~\ref{fig:lorenz}A,C). Figure~\ref{fig:lorenz}C depicts the link between the Lorenz's first dimension and the coefficients representation, by coloring the three most dominant operators' coefficients according to the Lorenz's first dimension values. One can see the clear change of colors in different areas of the $\bm{c}_t$ space. 

\looseness=-1

\subsection{Discrete-time dLDS identifies latent dynamics in \textit{C. elegans} data}\label{subsec:Celegans}

\begin{figure}[t]
    \centering
    \includegraphics[width=\textwidth]{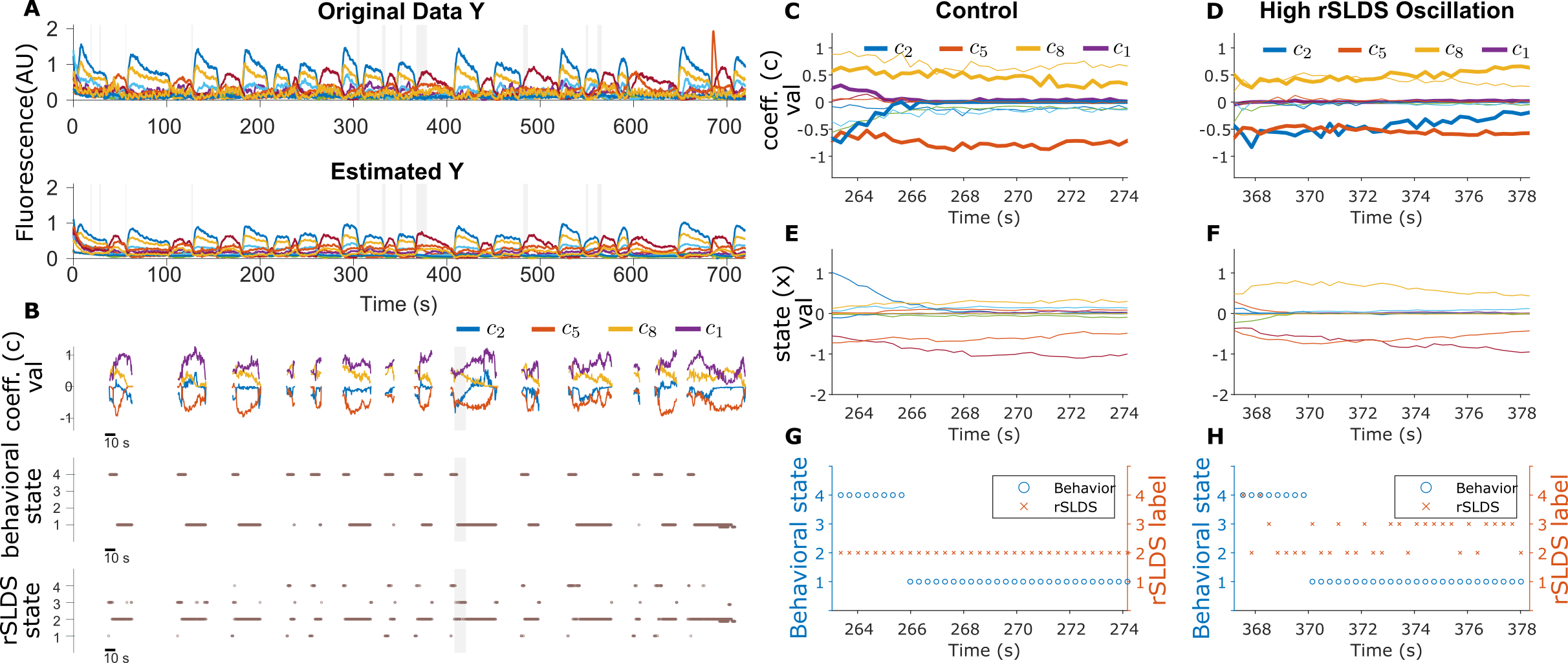}
    \caption{\textbf{Demixed dynamics in a single \emph{C. elegans} in the stimulated experimental setting, with 4 behavioral states.} \textbf{A:} Neuronal activity for a subset of neurons and their reconstruction ($R^2 = 0.74$) using dLDS with 10 DOs. 
    \textbf{B:} Regions corresponding to inferred behavioral states 4 and 1, with the top 4 dynamics coefficients for these epochs. Gray highlights indicate periods of high oscillation in the rSLDS state. By contrast, the dLDS dynamics coefficients during the 4-to-1 state transition varied more smoothly. $c_1$, $c_2$, $c_5$, and $c_8$ were highly active, with a quantitative change at the transition between states 4 and 1. \textbf{C-H:} A time interval where rSLDS exhibits state oscillations during a transition from state 4 to 1 (right column, \textbf{D, F, H}) juxtaposed with a control interval with stable rSLDS states during a similar transition (left column, \textbf{C, E, G}). Compared are the dLDS $\bm{c}$ traces (\textbf{C, D}), dLDS $\bm{x}$ traces (\textbf{E, F}), and Behavioral states (blue circles) labelled by Kato \emph{et al.} (2015) vs. rSLDS states (red crosses, not matched) (\textbf{G, H}).}
    \label{fig:celegans}
\end{figure}

Finally, we apply dLDS to ``whole brain'' \textit{C. elegans} calcium imaging recordings~\citep{Kato2015, Zimmer_2021, Linderman2019} (Fig.~\ref{fig:celegans}A). We benchmarked dLDS model against rSLDS in the experiments where~\cite{Kato2015} inferred the immobilized worms' pirouetting behavior under varying oxygen concentrations (four states: forward crawl, reverse crawl, sustained reverse crawl, and post-reversal turn) from the neural activity. dLDS revealed obscured differences in the neural dynamics during different behavioral states for further exploration. 

\begin{figure}
    \centering
    \includegraphics[width=\textwidth]{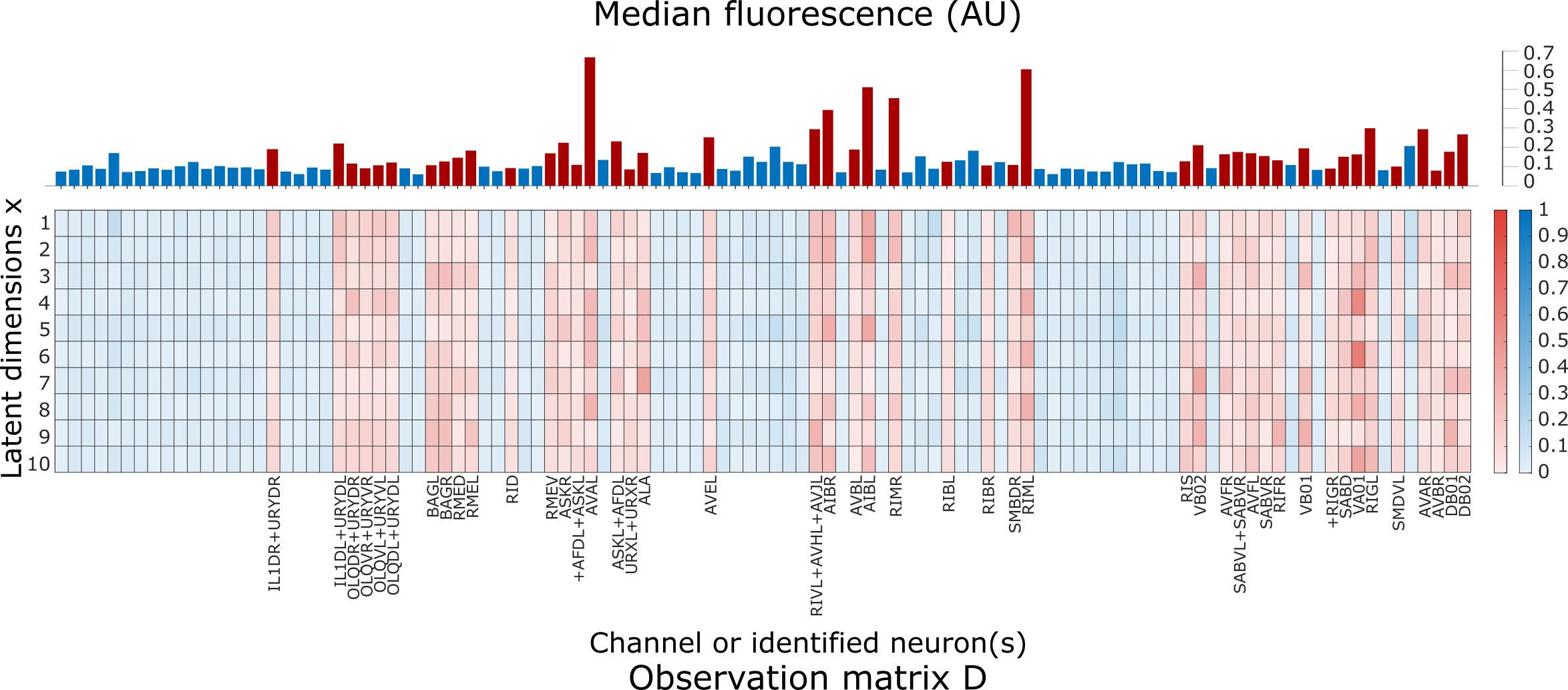}
\caption{\textbf{dLDS identifies important neurons in \textit{C. elegans} data via the observation matrix $\bm{D}$.} Top: Median fluorescence values for each recording channel (Worm 1, stimulated case). Channels labeled by neuron names represent individual neurons identified by Kato \emph{et al.}~\citep{Kato2015,Zimmer_2021}. Bottom: The observation matrix $\bm{D}$ mapping the latent states $\bm{x}$ to the fluorescence $\bm{y}$. }
    \label{fig:SupplCE}
\end{figure}

First, we observed that the map $\bm{D}$ from the latent space to the neural observations was able to highlight neurons that heavily overlapped with known neurons of interest and group them together in the latent space. For example, the stronger weights in $\bm{D}$ highlighted AVAL and RIML, interneurons involved in a backward motion, and VA01, a motor neuron (Fig.~\ref{fig:SupplCE}, neurons of interest in red). AVAL and RIML were both represented in latent dimensions 1, 2, 4, 6, and 8, whereas AIBL, an interneuron that instead promotes turns, was represented in latent dimensions 1, 2, 5, 8, and 10. VA01 was most represented in latent dimensions 3, 4, 6, 8, and 10, which suggests some shared utilization of this motor neuron for backward motion and some for turning. This mapping can be used in future studies to reconstruct measures of functional connectivity from latent space dynamics back to the neural activity (ambient) space. 

Second, we noticed obscured within-state evolving dynamics (Fig.~\ref{fig:celegans}B). This may indicate a gradual change in the worm's internal state or behavior in the middle of these discretely-labeled behavioral states; however, it is difficult to identify a behavioral or internal state correlated with the data provided. These results indicate that dLDS can be applicable to datasets with continuous descriptions of behavior as opposed to discrete behavioral state labels. 

Third, while rSLDS recapitulated 1-sparse behavioral state classification with high fidelity, we observed periods where the inferred discrete states oscillated unrealistically, e.g., between the post-reversal turn (rSLDS state 4) and forward crawling (rSLDS state 1). By contrast, dLDS adjusted the dynamics coefficients without changing the usual relationships between the dynamics. The traces maintained a standard motif shape and order of purple/yellow/blue/red from top to bottom, with blue activating during periods of high oscillation in the rSLDS classification (Fig.~\ref{fig:celegans}B). However, not every behavioral state transition from state 4 to state 1 (``4-to-1'') induced high oscillation in the rSLDS classification; we take as controls those time intervals where the rSLDS inferred states were stable, which we defined as a remaining constant for at least 10 time points in a row. Focusing on a specific example, the stable rSLDS control (Fig.~\ref{fig:celegans}C,E,G) and high rSLDS oscillation (Fig.~\ref{fig:celegans}D,F,H) intervals both appeared to reflect 4-to-1 behavioral state transitions according to the ground-truth inferred behavior labels from~\cite{Kato2015}. However, during the high oscillation interval, in the dLDS latent states, the yellow trace stayed on, unlike in the control interval (Fig.~\ref{fig:celegans}E,F). In the dLDS dynamics coefficients (Fig.~\ref{fig:celegans}C,D), several dynamics were active at the beginning of State 4 in the control interval but not in the high oscillation interval (264-266s vs. 367-370s). Moreover, in the high oscillation interval, the dynamic corresponding to the orange trace stayed active, unlike in the control interval. Both dLDS and rSLDS identified some differences in neural activity between these two time intervals. However, dLDS offers a finer resolution lens for parsing and interpreting the dynamics that create these differences.

\subsection{Additional experiments} We further tested dLDS in a higher-dimensional discrete-time setting (12x12 image patches), demonstrating the ability to recover ground truth dynamics in simple permutation tests, as well as identifying a decomposable transformation of image patches learned from natural videos (Appendix~\ref{sec:BBCsupp}).

\section{Discussion, Limitations, and Future Work}

In this work, we present a manifold-flow-inspired model of learning decomposed Linear Dynamical Systems. Our proposed model, dLDS, expands on the idea of switching linear systems to a model where linear combinations of a finite dictionary of systems can represent a richer set of dynamics. Unlike more unstructured dynamics models~\citep{harris2021time,proctor2016dynamic,luttinen2014linear}, the sparsity specifically provides an intuitive sense of interpretability for the dynamics that we observe over multiple examples. We present the model in both continuous- and discrete-time settings and provide both synthetic and real-data examples of learning the dynamics dictionary in practice.

Our model-learning provides an avenue by which we can estimate dynamical systems that are locally linear at each point, but whose parameters change over time. This enables both the estimation and tracking of non-stationary dynamics, as well as the approximation of nonlinear dynamics by treating the nonlinearity as a temporal non-stationarity. The learning procedure requires tuning a number of parameters, including regularization parameters and the number of latent and dynamics coefficients. Important targets of future work will be to develop approaches to automatically set the model hyperparameters. Furthermore, we aim to investigate additional aspects of the model such as its stability, its performance under different statistical assumptions (e.g. Poisson), and its predictive capability.

In this work, we primarily focused on the linear observation model (Eq.~\eqref{eq:1}) common in many neural data analysis methods. However, it is known that for manifold structured data, nonlinear models can often further reduce data dimensionality. Thus future methods should implement nonlinear observation models to reduce the dimensionality of the space in which the dynamics dictionary is learned, e.g., learning DOs in the latent space of an auto-encoder~\citep{Connor2021identitypreserving}.

The examples we present highlight a series of both simple and more complex systems. We note that even the simpler systems represent plausible system behaviors in real data that are not captured well by existing techniques. For example, the evolution of a system at different speeds indeed targets a weakness of switching models. However, it is not a contrived example. In fact, there are noted examples of neural systems that have warped timescales representing elongated processing of the same computation~\citep{Wang2017}.

One nuanced point that is critical to address is that our framework can model both nonlinear and non-stationary systems. In particular, dLDS achieves this ability, over models that purely target nonlinear systems~\citep{brunton2016discovering}, by modeling the nonlinearities as nonstationarities through the time-varying coefficients $\bm{c}_t$. The benefit of this model is that for nonlinear systems we expect, as we have seen in the FHN and Lorenz cases, regularities in the system behavior that can be tracked by analyzing the dynamics coefficients. Nonstationarities can also be identified in a similar way by noting non-periodic changes in $\bm{c}_t$ or changes in $\bm{c}_t$ in nearby locations in state space at different times. This nuance is important, as it is often of interest if the observed changes in neural dynamics are due to changes in the system behavior or due to the intrinsic nonlinear interactions. Further work should continue to try to identify markers to better disambiguate these cases. 

One direction that we have not explored in this work is the inclusion of a driving term in the dynamics equation, e.g., in discrete time by adding a term $\bm{B}\bm{u}_t$ such that $\bm{x}_t = \bm{F}_t\bm{x}_t + \bm{B}\bm{u}_t$. Here $\bm{u}_t$ can represent a set of control variables, and $\bm{B}$ is the projection of those inputs into the latent space. While this term is trivial to add to our basic framework (in fact, similar update rules can be included to learn $\bm{B}$), we focused here on the most basic case to demonstrate the utility of the dynamics model alone. Future work will focus on this term, in particular as it can be used to account for system inputs such as sensory stimuli in modeling neural dynamics.

\textbf{Limitations:} In addition to the above discussion points, other current limitations of our work are that under some parameter regimes, the estimates for the dynamics coefficients can be unstable. Additionally, we note that we observe better performance when normalizing and centering the datasets. These shortcomings recommend future directions for preventing severe instability in the inference step by building in bias terms, e.g., similar to the learned offset terms in other LDS approaches to approximating nonlinear systems~\citep{linderman2017bayesian}. We note that our model is not proposed to replace rSLDS, but to reveal how the activity of the same elements smoothly evolves over time, rather than a binary ‘switch’ between elements. The two methods will have different strengths depending on the assumed dynamics of a given system.


\section{Data and Code Availability}
The code for both the discrete and the continuous models can be found at https://github.com/dLDS-Decomposed-Linear-Dynamics. The discrete code can also be pip-installed using the \textbf{dLDS-discrete} Python package, as described in \url{https://pypi.org/project/dLDS-discrete-2022/}. Additionally, an interactive Python notebook of the discrete model visualization is available at \url{https://colab.research.google.com/drive/1PgskOtYoLL83ecXz_ALXk9oLofR2AeRf?usp=sharing}. 

Data from~\cite{Kato2015} were obtained from the Open Science Framework site~\citep{Zimmer_2021}.
The SSM Python package from the Linderman lab was used to run rSLDS~\citep{Linderman_SSM_Bayesian_Learning_2020}.

\section{Acknowledgments}

This work was supported in part by NSF grant CCF-1409422, James S. McDonnell Foundation grant number 220020399, the Georgia Institute of Technology, and Johns Hopkins University. N.M. was supported by the Kavli Foundation Kavli Discovery Award. Thank you to Sue Ann Koay for the original version of the manifold illustration in Figure~\ref{fig:introFig}. 



\newpage

\appendix
\setcounter{page}{1}

\section{Fitting as Variational EM}
\label{subsec:varEM}
We perform model fitting via a DL algorithm derived via a variational expectation-maximization (EM) approach, e.g.,~\citep{OLS:1996,dayan2001theoretical,barello2018sparse}. For shorthand notation, we consider here $Z = \{X,C\}$ (i.e., $\bm{z}_t = [\bm{x}_t^T, \bm{c}_t^T]^T$ the combined latent model coefficients. Furthermore each set variable $Y,Z$ represents the full dataset: $Y = \{\bm{y}_1,...,\bm{y}_t\}$ and $Z = \{\bm{z}_1,...,\bm{z}_t\}$. In the model parameter learning we wish to optimize the log-likelihood of the data given the parameters
\begin{gather}
    \log P(Y \mid \bm{D},\{\bm{f}_m\}). 
\end{gather}
Variational EM defines a surrogate distribution $Q(Z)$ as an approximation to the posteror $P(Z|Y)$ that enables the optimization of a tractable lower-bound on the log-likelihood. This is known as the evidence lower bound (ELBO):
\begin{align}
  \log P(Y \mid \bm{D},\{\bm{f}_m\})  &=  \log \int P(Y,Z \mid \bm{D},\{\bm{f}_m\}) \dif
                       Z = \log \int  Q(Z) \frac{P(Y,Z \mid \bm{D},\{\bm{f}_m\})}{Q(Z)}  \dif Z \nonumber \\
 &\geq \int  Q(Z)\log\left(\frac{P(Y,Z|\bm{D},\{\bm{f}_m\})}{Q(Z)}\right)\dif Z   \nonumber  \\
&= \log P(Y|\bm{D},\{\bm{f}_m\}) - D_{KL}\Big[Q(Z)\Big|\Big|P(Z|Y,\bm{D},\{\bm{f}_m\})\Big]
\nonumber \\
& \triangleq \mathcal{L}(Q,\bm{D},\{\bm{f}_m\}), \nonumber        
\end{align}
where the inequality above is due to Jensen's inequality, and $D_{KL}[Q(Z)||P(Z)]=\int Q(Z)$ $\log [ Q(Z) / P(Z)] dZ$ denotes the Kullback-Leibler (KL) divergence between $Q(Z)$ and $P(Z)$~\citep{bishop05,Blei16}.

Variational EM can thus be written as alternating coordinate ascent of the ELBO over the model coefficients and model parameters:
\begin{align}
    \mbox{E-step:} \qquad Q & \leftarrow  \arg\max_{Q}  \mathcal{L}(Q,\bm{D},\{\bm{f}_m\}) = \arg\min_{Q}  D_{KL}\Big[Q(Z)\Big|\Big|P(Z|Y,\bm{D},\{\bm{f}_m\})\Big]\nonumber \\
    \mbox{M-step:} \qquad \bm{D},\{\bm{f}_m\} & \leftarrow  \arg\max_{\bm{D},\{\bm{f}_m\}}  \mathcal{L}(Q,\bm{D},\{\bm{f}_m\}) = \arg \min_{\bm{D},\{\bm{f}_m\}}\; \mathbb{E}_Q \Big[ \log P(Y,Z|\bm{D},\{\bm{f}_m\}) \Big] . \nonumber
\end{align}

The simplest possible surrogate distribution $Q(Z)$ is a product of Dirac delta functions:
\begin{equation}
  Q(Z) = \prod_{t=1}^T \delta(\bm{z}_t-\bm{\gamma}_t),
\end{equation}
where $\{\bm{\gamma}_1, \ldots, \bm{\gamma}_T\}$ are the variational parameters governing $Q$~\citep{barello2018sparse}. As in traditional DL, we can write the E-step as
\begin{align}
D_{KL}\left[Q(Z)\Big|\Big|P(Z|Y,\bm{D},\{\bm{f}_m\})\right] &= \int Q(Z)[\log(Q(Z)) - \log(P(Z|Y,\bm{D},\{\bm{f}_m\}))]\nonumber\\
&= - P(\gamma|X,\bm{D},\{\bm{f}_m\}) + \int Q(Z)\log(Q(Z)),
\end{align}
where final term in the last line is vacuous and can be ignored. Thus the ELBO reduces to:
\begin{equation}
  \label{eq:7}
  \mathcal{L}(Q,\bm{D},\{\bm{f}_m\}) = \sum_{l} \int \delta(\bm{z}_t-\bm{\gamma}_t) \log P(\bm{y}_t, 
  \bm{z}_t | \bm{D},\{\bm{f}_m\}) \dif z_t = \sum_l \log P(\bm{y}_t, \bm{\gamma}_t | \bm{D},\{\bm{f}_m\}),
\end{equation}
the $\arg\max$ of which, with respect to $\bm{\gamma}_t$, coincides with the MAP estimate over the model coefficients. Following the E-step, the M-step updates the model parameters $\bm{D},\{\bm{f}_m\}$ to further minimize the ELBO. Rather than fully optimize given sample $\bm{\gamma}$, we follow~\citep{OLS:1996} and instead take a single gradient step at each iteration. With small samples, this algorithm each gradient step is noisy, mimicking a stochastic gradient descent version of variational EM~\citep{Neal98}.


\section{Duffing Dynamics and Linearization Components Recovery}

As an example of the application of our model's abilities, we demonstrate its capability in recovering the latent components of a linearization of the Duffing oscillator. The Duffing oscillator is a classical example of a nonlinear, second-order differential equation that exhibits chaotic behavior. It is described by the following equation:

\begin{equation}
\ddot{x} + \delta \dot{x} + \alpha x + \beta x^3 = \gamma \cos(\omega t),
\end{equation}
where $\alpha$, $\beta$, and $\gamma$ are parameters that control the behavior of the oscillator, and $\gamma \cos(\omega t)$ represents a periodic driving force with amplitude $\gamma$ and frequency $\omega$. 

To apply our model to the Duffing oscillator, we define $y = \dot{x}$. We can then rewrite the Duffing oscillator equation in terms of $y$ as follows:
\begin{equation}
\dot{y} + \delta y + \alpha x + \beta x^3 = \gamma \cos(\omega t).
\end{equation}

We can represent the Duffing oscillator using a decomposed linear dynamical systems model by discretizing the system over time. We use the state vector $[x_t, y_t]$ to represent the state of the system at time $t$. The discrete-time model of the Duffing oscillator is given by the following equation:

\begin{equation}
\begin{bmatrix}
\dot{x}_t \\
\dot{y}_t
\end{bmatrix} = \begin{bmatrix}
0 & 1 \\
-\alpha - \beta x_t^2 & -\delta
\end{bmatrix} \begin{bmatrix}
x_t \\
y_t
\end{bmatrix} + \begin{bmatrix}
0 \\
\gamma \cos(\omega t)
\end{bmatrix}.
\end{equation}

We can then apply our model to this discrete-time Duffing oscillator equation. We estimate the transition matrix $A_t$ by fitting our model to the time series data. We can decompose $A_t$ into two matrices, $F_t = F_{t,1} + F_{t,2}$, where $F_{t,1}$ and $F_{t,2}$ correspond to distinct physical processes.

Specifically, the matrix $F_{t,1} = \begin{bmatrix}
1 & \Delta t \\
-\alpha \Delta t & 1 - \delta \Delta t
\end{bmatrix}$ describes the linear dynamics of the system, while $F_{t,2} = \begin{bmatrix}
0 & 0 \\
-\beta x_t^2 \Delta t & 0
\end{bmatrix}$ describes the nonlinearity of the system.

Using these matrices, we can write the discrete-time Duffing oscillator equation in matrix multiplication form as follows:

\begin{equation}
\begin{bmatrix}
x_{t+1} \\
y_{t+1}
\end{bmatrix} = \left(\begin{bmatrix}
1 & \Delta t \\
-\alpha \Delta t & 1
\end{bmatrix} + \begin{bmatrix}
0 & 0 \\
-\beta x_t^2 \Delta t & 0
\end{bmatrix}\right) \begin{bmatrix}
x_t \\
y_t
\end{bmatrix} + \begin{bmatrix}
0 \\
-\cos(t)
\end{bmatrix},
\end{equation}

where $\cos(t)$ represents the cosine function evaluated at time $t$. By estimating the matrices $F_{t,1}$ and $F_{t,2}$ using our model, we can recover the underlying linear and nonlinear components of the Duffing oscillator dynamics. By applying dLDS to the Duffing oscillator, we highlight the model's capacity to reconstruct nonlinear dynamics using basic linear components whose coefficients change over time and to recover the ground truth basic linear elements that underlie these complex dynamics. These results emphasize the potential of dLDS for extracting valuable insights from intricate systems, by facilitating the reconstruction of nonlinear dynamics with basic linear components, and by retrieving the underlying basic linear elements.

\begin{figure}[t!]
    \centering
    \includegraphics[width=1\textwidth]{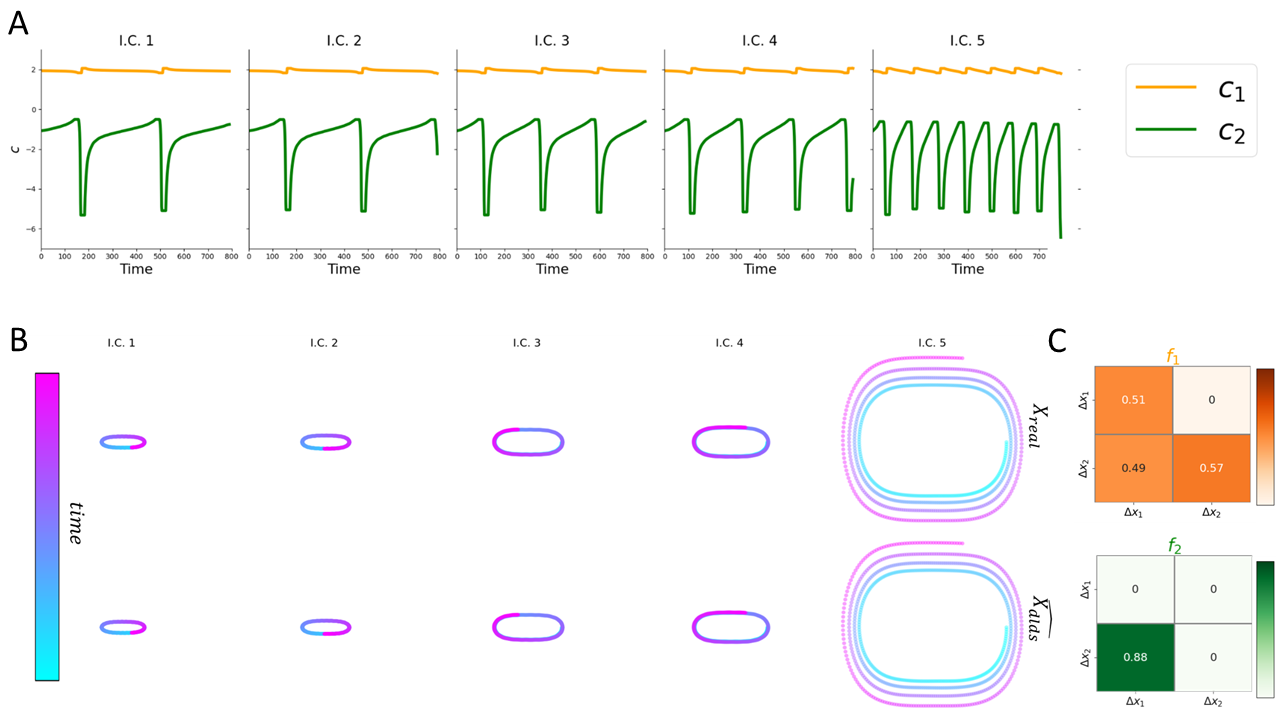}
    \caption{\textbf{Results of dLDS analysis on the Duffing oscillator.} A) Time traces (orange and green) of each dynamic operators (DO) for different initial conditions (I.C.) in separate subplots. B) Comparison between the real dynamics of the Duffing oscillator and the reconstructed dynamics by dLDS for each I.C. The color represents time. C) Heatmaps of each DO, where the first DO shows the main diagonal and other non-zero elements, as its coefficient in A is 2, while the second DO shows the lower left diagonal element and all other elements are zero.}
    \label{fig:duffing}
\end{figure}

\section{Continuous-time dLDS learning}

For completeness, we provide here in Algorithm~\ref{alg:ctmt} the dLDS learning algorithm for continuous time dynamics. In this work for continuous time we assume that $\bm{D} = \bm{I}$.

\begin{algorithm}[t!]
\caption{Continuous-time dLDS training assuming $\bm{D}=\bm{I}$}\label{alg:ctmt}
\begin{algorithmic}
\State \textbf{Input} $Y$, $\lambda_1$, $\lambda_2$, $\eta$, $L$ \Comment{Data, penalty on $c$, penalty on $G$, learning rate, n dictionary elements}
\State \textbf{Initialize}  $\{\bm{G}_i\}_{l=1:L}, D$  \Comment{Randomly initialize sub-dynamics, observation model} 
\State Normalize each dynamical system by dividing by its spectral radius 

\While{not converged} \Comment{Iterating until convergence}
    \For{$t$ in $\{0,\dots,T-1$\}}  \Comment{Run for every time-adjacent data pair}
        \State Randomly sample $\{\hat{c}_l(t)\}_{l = 1}^L$
        \State Update $\{{\hat{c}_l(t)}\}_{l = 1}^L \gets \arg\min_{\{c_l(t)\}} E$ 
        \State Update $\hat{G} \gets \arg\min_G E$  
    \EndFor
\EndWhile
\end{algorithmic}
\end{algorithm}

\section{Special case of no observation model}
\label{sec:NoObs}
For completeness, we provide here in Algorithm~\ref{alg:cap3} the dLDS learning algorithm for discrete-time dynamics under the condition that $\bm{D} = \bm{I}$. This special case is less computationally intensive and enables the learning of dynamics in the native data space. 

\begin{algorithm}[t!]
\caption{Dynamics dictionary learning for $\bm{D}=\bm{I}$}\label{alg:cap3}
\begin{algorithmic}
\State \textbf{Input} $Y$, $\lambda_1$, $\eta$, $M$ \Comment{Input observations and hyperparameters}
\State \textbf{Initialize}  $\{\bm{f}_m\}_{m=1:M}$, \Comment{Initialize dynamics dictionary randomly} 
\State Normalize each $\bm{f}_m$ to unit spectral radius
\While{not converged} \Comment{Iterating until convergence}
    \State $\widehat{\bm{c}}_t = \arg\min_{\bm{c}} \|\bm{y}_t - \sum_m\bm{f}_mc_{mt}\bm{y}_{t-1}\| + \lambda_1\|\bm{c}\|_1$
        \State Update each $\bm{f}_m$ via Equation~\ref{eqn:Fupdate} 
        \State Normalize each $\bm{f}_m$ to unit operator norm
    \If{$rMSE$ does not change} \Comment{Check if the algorithm stuck in a local minimum} 
            \State ${f_{m}} \gets{f_{m} +\nu}$  \Comment{Add random noise to each $\{f_{m}\}$} 
    \EndIf
\EndWhile
\end{algorithmic}
\end{algorithm}

\section{A note on LASSO solvers}
We found that the correlations between $\bm{f}_m\widehat{\bm{x}}_{t}$ could be large for certain time points. These correlations meant that some $\ell_1$ regularize least-squares solvers would exhibit instability during learning. We noticed this in particular for the primary LASSO functions in both MATLAB and Python. We found that instead, the SPGL1 solver of the pylops package~\citep{ravasi2020pylops} was more robust in Python, and the TFOCS software~\footnote{https://github.com/cvxr/TFOCS}~\citep{becker2011templates} was more robust in MATLAB. While TFOCS solves the LASSO program directly, SPGL1 solves a slightly modified version:

\begin{equation} \label{eq:lasso_spgl1}
\ {\hat{c_t} = \arg\min_{c}{{\left\|x_{t+1} - \sum_{m=1}^M \bm{f}_m c_{m}(t) * x_t\right\|_2} \qquad \textrm{s.t.} \qquad {||c_{t}||_1} \leq \tau}}.
\end{equation} 

Importantly, despite presenting only the SPGL1 results in the above paper, the Python code enables the user to choose from a wide range of solvers (including FISTA, ISTA, Sklearn LASSO~\citep{scikit-learn}, OMP), and the decision for which solver to use, is up to the user, and should depend on the data properties and the user's goals in running the model.

\section{Additional information on experiments}
\subsection{Continuous-time model parameters}

Continuous-time models were implemented in PyTorch. Model parameters for each experiment are shown in Table~\ref{table:modelParameters}.

\begin{table}[t!]
\centering
\begin{tabular}{c | c c c c c c} 
 Experiment & $L$ & $\lambda_G$ & $\lambda_c$ & $\eta_G$ & $\eta_c$ & $\gamma$ \\ [0.5ex] 
 \hline
 Speed & 4& 1 & 1e-1 & 1e-1 & 1e-2& 0.985\\ 
 Rotation & 4& 20 & 8e-2 & 5e-3 & 1e-2 & 0.985 \\

\end{tabular}
\caption{\textbf{Continuous-time experiment model parameters.} $L$ is the total number of dictionary elements initialized. $\lambda_G$ is regularization placed on dictionary elements; we control this value through the weight decay. $\lambda_c$ is the regularization strength placed on the coefficients. $\eta_G$ and $\eta_c$ are the learning rates for dictionary elements and coefficients, respectively. $\gamma$ is the rate for the learning rate decay schedule.}
\label{table:modelParameters}
\end{table}

\subsection{FitzHugh-Nagumo (discrete-time model)}
We used the Python discrete code for the FHN case. The iterative model ran until convergence (reconstructed error < 1e-8) or until 
reaching a maximum of 6,000 iterations. The ground truth for the FHN dynamics was created based on ~\eqref{eq:FHN}, using 1000 samples with time intervals of 0.2  (s.t. $t_{\max}= 200$).
We used $M=2$ dictionary elements, an initial value of $\eta=30$ (from  ~\eqref{eqn:Fupdate}), while its decay rate over the training iterations was set to $\gamma=0.99$; The standard-deviation of the perturbations added randomly to each {$\bm{f}_m$} in case of local minimum, as described in Algorithm~\ref{alg:cap}, was set to 0.1.

For the regularized dynamics case, the hyperparameters of the SPGLl solver, {"iter lim"}, the maximum number of solver iterations in each coefficients updating step, was set to 10, and $\tau$ (from~\eqref{eq:lasso_spgl1}) was set to 0.3. 

For the unregularized case, the following pseudo-inverse was used for each the updating step of the coefficients in each time point and in each iteration:
\begin{eqnarray} 
\widetilde{\bm{F_t}} & = & \left[\bm{f}_1 \bm{x}_{t}, \bm{f}_2 \bm{x}_{t}, ..., \bm{f}_M \bm{x}_{t} \right] \in \mathbb{R}^{(2 \times M)} \nonumber \\
\bm{x}_{t+1}  & = &  \widetilde{F_t}\bm{c}_t \nonumber \\
\widehat{\bm{c}_t} & = & {\widetilde{\bm{F_t}}^\dagger} \bm{x}_{t+1},
\label{eq:lasso_pseudo}
\end{eqnarray} 
where $\dagger$ denotes the pseudo-inverse. 

\subsection{The Lorenz attractor (discrete-time model)}

We used the Python discrete code for the Lorenz attractor case, with different options for the number of dictionary elements. In the paper, the results of our model for $M = 5$ dictionary elements are presented, along the corresponding results from the reference  (rSLDS with 5 discrete states). 

As in the FHN case, the dLDS iterative model ran until convergence (reconstructed error < 1e-8) or until 
reaching a maximum of 6,000 iterations. The ground truth for the Lorenz attractor was created based on Equation~\eqref{eq:LORENZ}, using 1000 samples with time intervals of 0.01 (s.t. $t_{\max} = 10$).
Similarly to the FHN, the initial value of $\eta$ was set to 30, while its decay rate over the training iterations was set to 0.99. The standard-deviation of the perturbations added randomly to each $f_i$ in case of local minimum, as described in Algorithm~\ref{alg:cap}, was set to 0.1.

The updating step of the coefficients was performed similarly to the updating methodology presented above for the FHN case, according to which in the unregularized case, the coefficients were updated using the pseudo-inverse, and for the regularized case, the coefficients were updated using the SPGL1 solver. 

With respect to the hyperparameters of the SPGLl solver in the Lorenz case, {"iter lim"} was set to 10, and {$\tau$} (from Equation~\eqref{eq:lasso_spgl1}) was set to 0.55. A comparison of model performance measures between dLDS and rSLDS is shown in Table~\ref{table:summaryResults}.

\subsection{Summary of model performance comparisons}

Table \ref{table:summaryResults} compares rSLDS and dLDS summary statistics for each experiment. 

\begin{table}[t!]
\centering
\begin{tabular}{c cc c cc} 
& \multicolumn{2}{c}{rSLDS} && \multicolumn{2}{c}{dLDS}\\
\cmidrule(lr){2-3} \cmidrule(lr){5-6}
 Experiment & $R$ & $R^2$ &&  $R$ &  $R^2$ \\ [0.5ex] 
 \hline
  Synthetic speed & 0.99  & 0.99 && 0.99 & 0.99 \\
  Synthetic rotation & 0.99  & 0.99  && 0.99  & 0.99  \\
  Unregularized FHN & 1.0  & 1.0  && 1.0 & 1.0   \\
  Regularized FHN & -  & -  && 1.0 & 1.0   \\
 Unregularized Lorenz & 0.99 & 0.99 && 1.0 & 1.0 \\
 Regularized Lorenz & -  & -  && 0.93 & 0.70\\
 \emph{C. elegans} &  0.96 & 0.92 && 0.86 & 0.74 \\
 
\end{tabular}
\caption{\textbf{Pearson  correlation and $R^2$ values for one-step prediction for rSLDS vs dLDS, for each experiment.}}
\label{table:summaryResults}
\end{table}

\section{Comparison between dLDS and rSLDS for the FHN model}

\begin{figure}[t]
    \centering
    \includegraphics[width=0.8\textwidth]{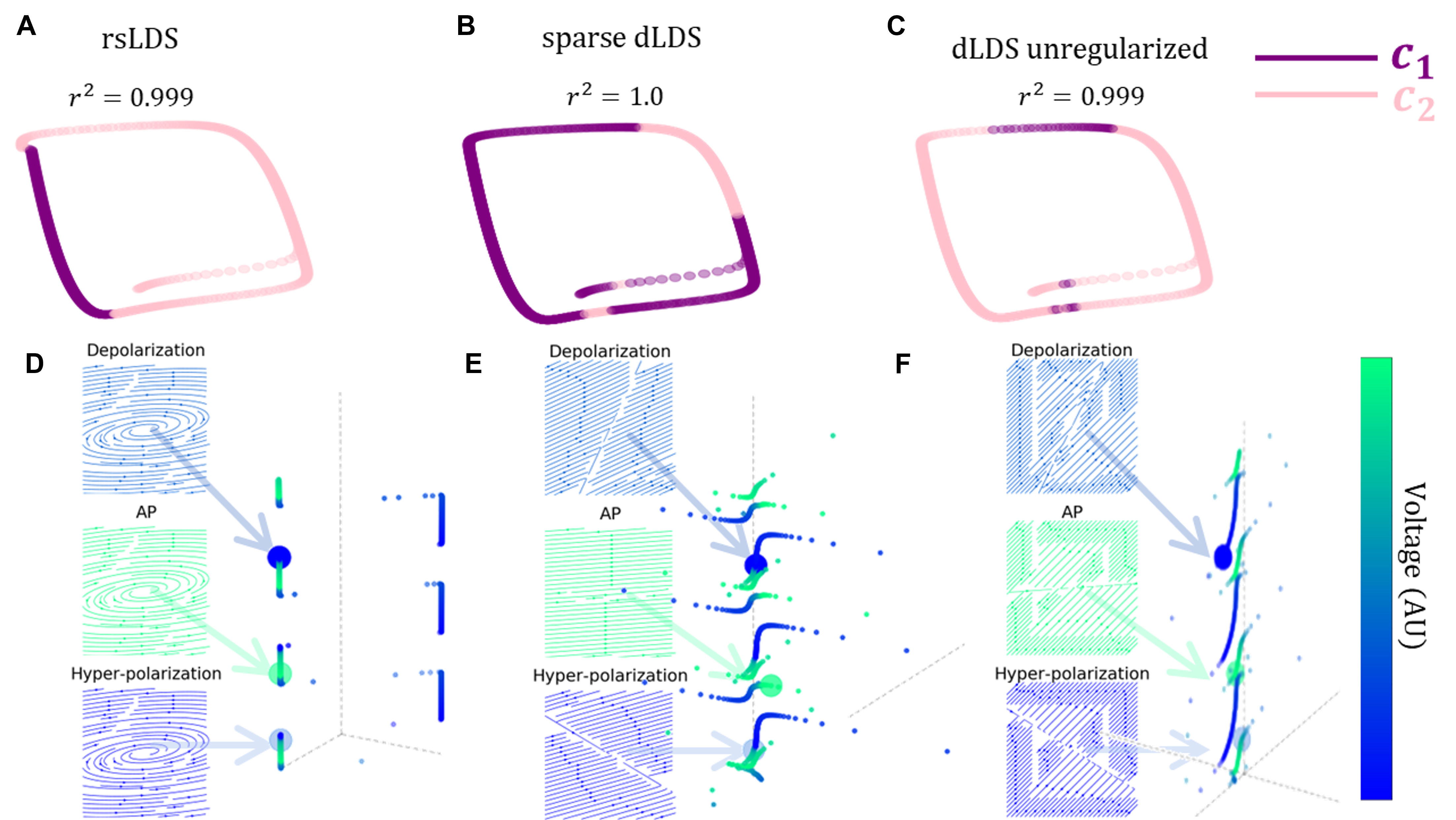}
    \caption{
    \textbf{Learned representations using dLDS and rSLDS for the FHN oscillator.} Note three time points of interest: repolarization, action potential peak, and hyperpolarization. With $M = 2$, dLDS can reconstruct the three distinct states, while rSLDS can only capture two reconstructed states in this case.
     A, B, C: All three models were able to reconstruct the FHN dynamics.
     D: The coefficients obtained by the rSLDS are restricted to the axes, resulting in no more than two distinct reconstructed states to describe the action potential cycle.
    E: Although the coefficients of the regularized dLDS tend to live on the axes due to the sparse regularization, this constraint is softer than of the rSLDS (in which living outside the axes is not possible). 
    F: Coefficients space obtained for the unregularized dLDS model. Most coefficients do not necessarily live on the axes, since no regularization was applied.}
    \label{fig:FHNCompareTorSLDS}
\end{figure}

In the main text we focus on comparing dLDS for the FHN oscillator under regularized and unregularized conditions to emphasize the role of regularization over the dynamics coefficients $\bm{c}_t$. Here we further demonstrate the comparison between dLDS to rSLDS (Fig.~\ref{fig:FHNCompareTorSLDS}). 

In contrast to the rSLDS model, for which the coefficients are binary, in our model the coefficients can take on continuous values. Hence, contrasting the observed coefficients-space spanned by rSLDS (\ref{fig:FHNCompareTorSLDS}D) and dLDS (\ref{fig:FHNCompareTorSLDS}E,F), in dLDS the dynamics representations are not limited to discrete locations on the axes, but can travel along them, resulting in a more flexible representation without the need to increase the number of dynamical systems learned. Specifically, we identify that while rSLDS learns slightly varying dynamical systems, while dLDS learns reorientations to different axes which more smoothly trade off with each other as the system rotates about the attractor. 

Additionally, as the $\ell_1$ regularization over $\bm{c}_t$ in dLDS increases, the coefficients become more similar to those obtained by rSDLS, namely, more restricted to the axes. Thus, modulating the regularization in our model makes possible the creation and exploration of a continuum of representations whose coefficients-space range from switched systems (high regularization) to arbitrary structured (unregularized), as described in Figure~\ref{fig:introFig}B. 

\section{Sparse video example}

\begin{figure}
    \centering
    \includegraphics[width=0.8\textwidth]{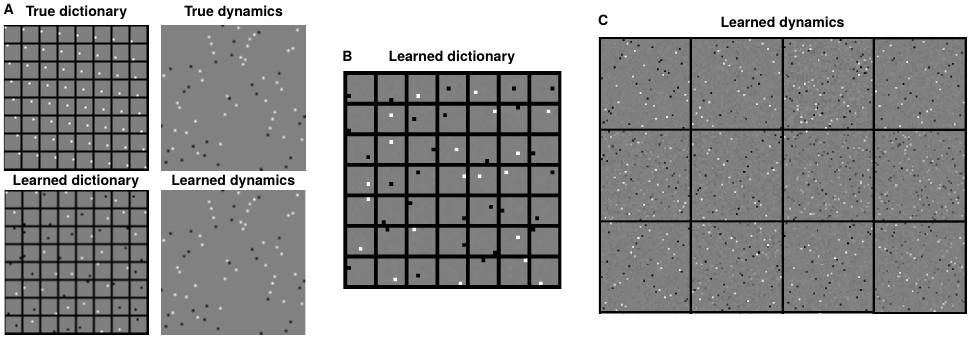}
    \caption{\textbf{Pixel permutation example.} A: Example test set consisting of a sparse number of pixels being permuted via an unknown permutation matrix. dLDS recovers in this case both the pixel-sparse dictionary as well as the ground truth permutation matrix. B: For a more complex example where multiple permutation matrices may be used (sometimes in tandem to split or merge pixels), the correct pixel-sparse dictionary is again learned. C: For the same example as B, the set of permutation matrices is learned, capturing the underlying dynamics.}
    \label{fig:permute}
\end{figure}

To test the model in a sparse higher-dimensional setting we simulate a single dynamics function is present ($M=1$) and the sparsity dictionary as the canonical basis ($\bm{D} = \bm{I}$). This test will check if our algorithm can accomplish simple system identification as a special case. We modeled the single dynamics function as a permutation matrix concatenated with a scaling matrix, i.e., signal coefficients move around and may be scaled (Fig.~\ref{fig:permute}A). The learned and true models are a very close match, differing by only a permutation and sign change (the same ambiguity present in all DL methods). 

In a more complex simulation, we simulate a dictionary of twelve distinct scaled permutation functions, only two of which are used at any time step (i.e. the sparsity of $\bm{c}_t$ is two). 
This system induces complex, highly non-stationary dynamics. 
Figure~\ref{fig:permute}B,C depicts the results of the learning procedure, demonstrating that the sparsity dictionary is again learned up to a permutation and sign change, and the learned dynamics functions are again close matches to the true dynamics (i.e., we recover 12 scaled permutation matrices).

\section{BBC video example} 
\label{sec:BBCsupp}

\begin{figure}[t]
    \centering
    \includegraphics[width=\textwidth]{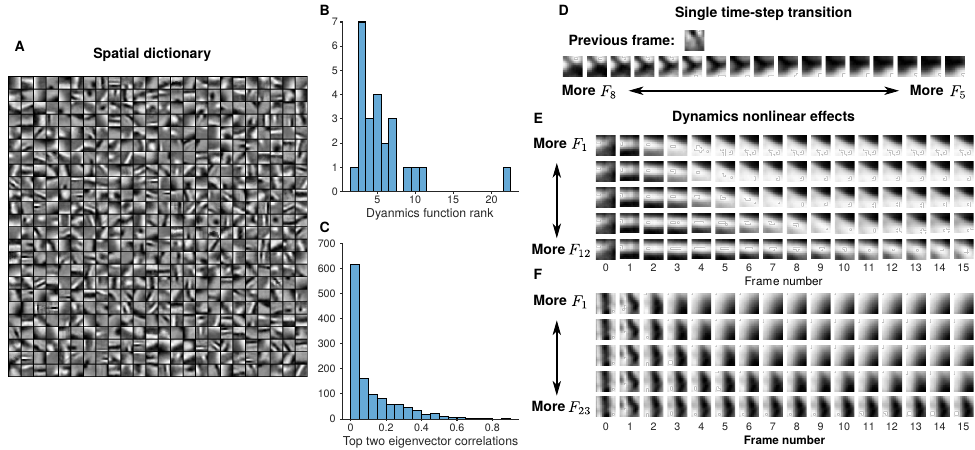}
    \caption{\textbf{Results of dynamics learning on natural image patches.} A: The spatial dictionary interestingly retains the Gabor-like structure seen in previous static DL algorithms. 
    B: Learned dynamics are low rank. The typical rank of each $\bm{f}_m$ (aside from one almost full-rank function) ranges from 2-10. C: the correlations between the top two eigenvectors show that the dynamics are mostly un-aligned, yet overlap, allowing for second-order effects when combining dynamics. 
    D: Linear combinations of dictionary elements can achieve nonlinear effects. Starting from the same frame, the next frame changes continuously between two possible next frames as the fraction of each dynamics function used is swept from completely using $\bm{f}_8$ to $\bm{f}_5$. 
    E,F: Examples of dynamics combinations that achieve nonlinear effects. For each of changing from using more of $\bm{f}_1$ and $\bm{f}_{12}$ and $\bm{f}_1$ and $\bm{F}_{23}$, the overall effect (rotation/expansion and outward expansion respectively) happens with faster or slower speeds. }
    \label{fig:BBC}
\end{figure}

To test dLDS on higher-dimensional real data, we learn a dynamics dictionary for natural video sequences. For computational considerations we restricted our algorithm to learn representations of 12x12 pixel patches, and learned a 4x overcomplete sparsity dictionary concurrently with 25 576x576 DOs. As no ground truth is available for video sequences, we instead qualitatively explore learned dictionaries. First we note that the sparsity dictionary recovered the expected Gabor-like statistics for image patches~\citep{OLS:1996,aharon2006rm} (Fig.~\ref{fig:BBC}A). This result matches the intuition that the spatial statistics are not qualitatively changed by including the temporal model. To assess the dynamics we note that despite the high-dimensional nature of the data, the learned dynamics were relatively low-dimensional (rank 2-10), with one exception that had almost full rank (Fig.~\ref{fig:BBC}B). Additionally, the top eigenvectors tend to be correlated, but not overly so. The correlations cluster around $\sim0.2$ with some correlations as high as 0.8 (Fig.~\ref{fig:BBC}C). This indicates that the learned functions are neither independent nor identical. Thus, interactions between the dLDS DOs permit flexible nonlinear behaviors. 

Three such examples are shown in Figure~\ref{fig:BBC}D-F. First we project a single frame forward by a combination of two overlapping dynamics, $\bm{f}_1$ and $\bm{f}_5$. As the weight on $\bm{f}_1$ is reduced and that on $\bm{f}_5$ increased, the projection changes from exaggerating the linear feature in the top-left to inverting the image with an emphasis on the bottom right. Similar effects appear in iterated dynamics projections, for example changing the weights on $\bm{f}_1$ and $\bm{f}_{12}$. This combination effectively rotates a bar over time, and the speed of rotation depends on the amount of $\bm{f}_1$ vs. $\bm{f}_{12}$ in the linear combination. This type of speed modulation is impossible in a switched model unless one mode for each speed is included, which is untenable for a continuum of speeds. Similarly when $\bm{f}_1$ and $\bm{f}_{23}$ are traded off, a vertical bar slowly has the bottom edge expand to encompass the bottom-right corner, again with different speeds depending on the ratio chosen. 

\section{Invariance of the model to transformations in the latent state}

Consider the base model complete with the observation equation and decomposed dynamics,
\begin{gather}
    \bm{y}_t = \bm{D}\bm{x}_t, \qquad\qquad
    \bm{x}_t11 = \left[ \sum_{m=1}^M \bm{f}_m c_{mt}  \right]\bm{x}_{t-1}.
\end{gather}

For any learned model, we can always define a transformation of the latent space via an invertable matrix $\bm{U}$ such that
\begin{gather}
    \bm{z}_t = \bm{U}^{-1}\bm{x}_t  \qquad  \bm{x}_t = \bm{U}\bm{z}_t. 
\end{gather}

This transformation results in an equivalent solution
\begin{gather}
    \bm{y}_t = \bm{D}\bm{U}\bm{z}_t, \qquad\qquad     \bm{z}_t  = \left[ \bm{U}^{-1}\sum_{m=1}^M \bm{f}_m\bm{U} c_{mt}  \right]\bm{z}_{t-1}, 
\end{gather}
i.e., an equivalent set of parameters $\widetilde{\bm{D}}=\bm{D}\bm{U}$ and $\widetilde{f}_m = \bm{U}^{-1}\sum_{m=1}^M \bm{f}_m\bm{U}$ result in the same sequence of dynamics but in a transformed latent space. One way to prevent the rotational ambiguity is to assume structure over the latent space, such as we implement via sparsity over $\bm{x}_t$, which enables us to learn the correct representation (up to a permutation and sign-flip) of observation model (Fig.~\ref{fig:permute}).

\end{document}